%% file: main_ieee_tran.tex
\documentclass[compsoc, cspaper]{IEEEtran} 
\usepackage{array}
\usepackage{stfloats}
\usepackage{verbatim}
\usepackage{cite}
\usepackage{amsmath,amssymb,amsfonts}
\usepackage{algorithmic}
\usepackage{graphicx}
\usepackage{textcomp}
\usepackage{multirow}
\usepackage{booktabs,tabularx,makecell}
\usepackage[inline]{enumitem}
\usepackage[table, xcdraw]{xcolor}
\usepackage{scalerel}
\usepackage{tikz}
\usepackage{xcolor}

\usetikzlibrary{svg.path}
\definecolor{orcidlogocol}{HTML}{A6CE39}
\tikzset{
  orcidlogo/.pic={
    \fill[orcidlogocol] svg{M256,128c0,70.7-57.3,128-128,128C57.3,256,0,198.7,0,128C0,57.3,57.3,0,128,0C198.7,0,256,57.3,256,128z};
    \fill[white] svg{M86.3,186.2H70.9V79.1h15.4v48.4V186.2z}
                 svg{M108.9,79.1h41.6c39.6,0,57,28.3,57,53.6c0,27.5-21.5,53.6-56.8,53.6h-41.8V79.1z M124.3,172.4h24.5c34.9,0,42.9-26.5,42.9-39.7c0-21.5-13.7-39.7-43.7-39.7h-23.7V172.4z}
                svg{M88.7,56.8c0,5.5-4.5,10.1-10.1,10.1c-5.6,0-10.1-4.6-10.1-10.1c0-5.6,4.5-10.1,10.1-10.1C84.2,46.7,88.7,51.3,88.7,56.8z};
 }
}
\newcommand\orcidicon[1]{\href{https://orcid.org/#1}{\mbox{\scalerel*{
\begin{tikzpicture}[yscale=-1,transform shape]
\pic{orcidlogo};
\end{tikzpicture}
}{|}}}}

\usepackage{hyperref} 
\usepackage{cleveref}

\def\BibTeX{{\rm B\kern-.05em{\sc i\kern-.025em b}\kern-.08em
    T\kern-.1667em\lower.7ex\hbox{E}\kern-.125emX}}
\usepackage{balance}
\begin{document}
\title{FOOL: Addressing the Downlink Bottleneck in Satellite Computing with Neural Feature Compression}
\author{Alireza Furutanpey
\orcidicon{0000-0001-5621-7899}, 
Qiyang Zhang
\orcidicon{0000-0001-7245-1298},
Philipp Raith
\orcidicon{0000-0003-3293-9437}, 
Tobias Pfandzelter
\orcidicon{0000-0002-7868-8613}, 
\\ Shangguang Wang \orcidicon{0000-0001-7245-1298}, 
\textit{Senior Member, IEEE},
Schahram Dustdar
\orcidicon{0000-0001-6872-8821}, 
\textit{Fellow, IEEE}}

\IEEEtitleabstractindextext{%
 \begin{abstract}
 \input{abstract}
\end{abstract}
}
\maketitle
\IEEEdisplaynontitleabstractindextext

\input{sections/introduction}
\input{sections/related_work}
\input{sections/motivation_background}
\input{sections/method_systems}
\input{sections/method_compression}
\input{sections/evaluation/evaluation}
\input{sections/limitations}
\input{sections/conclusion}
\section*{Acknowledgment}
We thank Alexander Knoll for providing us with the hardware infrastructure. Kerstin Bunte for her valuable suggestions. Florian Kowarsch for the fruitful discussions we held. The authors acknowledge TU Wien Bibliothek for financial support through its Open Access Funding Programme.
\bibliographystyle{ieeetr}
\bibliography{main}
\input{biography/biography}
\end{document}

%% file: abstract.tex
Nanosatellite constellations equipped with sensors capturing large geographic regions provide unprecedented opportunities for Earth observation.
As constellation sizes increase, network contention poses a downlink bottleneck. Orbital Edge Computing (OEC) leverages limited onboard compute resources to reduce transfer costs by processing the raw captures at the source. However, current solutions have limited practicability due to reliance on crude filtering methods or over-prioritizing particular downstream tasks. 

This work presents an OEC-native and task-agnostic feature compression method that preserves prediction performance and partitions high-resolution satellite imagery to maximize throughput. Further, it embeds context and leverages inter-tile dependencies to lower transfer costs with negligible overhead. While the encoding prioritizes features for downstream tasks, we can reliably recover images with competitive scores on quality measures at lower bitrates. We extensively evaluate transfer cost reduction by including the peculiarity of intermittently available network connections in low earth orbit. Lastly, we test the feasibility of our system for standardized nanosatellite form factors. We demonstrate that the proposed approach permits downlinking over 100$\times$ the data volume without relying on prior information on the downstream tasks.


%% file: sections/introduction.tex
\section{Introduction} \label{sec:introduction}
\IEEEPARstart{T}{he} development of commercial ground stations~\cite{commercialgroundstations} and the advancement in aerospace technology has enabled the emergence of nanosatellite constellations~\cite{nanoconstellation} in low earth orbit (LEO) as a novel mobile platform. 
The standardization of small form factors, such as CubeSat~\cite{lee2009cubesat}, reduces launch costs, allowing for frequent updates and deployments. Manufacturers typically equip satellites with sensors to capture large geographic regions. The downlinked satellite imagery enables Earth observation (EO) services with socially beneficial applications,
such as agriculture~\cite{agriculture} and disaster warning~\cite{disasterwarning}.
Nonetheless, most constellations follow a ``bent pipe’' architecture where satellites downlink raw sensor data for processing in terrestrial data centers. Notably, given the constraints of orbital dynamics, satellites may only establish a connection for a few minutes. For example, the Dove High-Speed Downlink (HSD) system~\cite{dovehsd} provides segments with volumes as low as 12 GB during a single ground station pass.

As constellation sizes and sensor resolutions increase, downlink bandwidth cannot keep up with the accumulating data volume~\cite{vasisht2021l2d2,tao2023transmitting}. Additional ground station equipment may prevent link saturation. However, building and maintaining them, including licensing the necessary frequencies, is a significant cost factor for satellite operation. As an alternative, Orbital Edge Computing (OEC) proposes processing data at the source~\cite{denby2020,furano2020towards,wu2023comprehensive, Giuffrida2022phi, scompsurvey}.
Recent work on reducing bandwidth requirements in OEC is roughly categorizable in aggressive (task-oriented) filtering and compression~\cite{oecsurvey2023}. The former relies on subjective value measures that restrict their practicability to coarse-grained tasks, such as de-duplication or cloud filtering. The latter constrains entire missions to particular tasks or prediction models. We argue that the limitations of existing compression or other data reduction approaches are particularly adverse to OEC. 

First, the CubeSat design is intended for short-duration missions~\cite{lee2009cubesat} (typically up to 3-5 years), and despite waning prices, launching sensor networks in space is still associated with substantial logistical, administrative, and monetary costs. Therefore, it seems undesirable to designate entire constellations to a small subset of tasks and Deep Neural Network (DNN) architectures. 
More pressingly, irrespective of whether current codecs can prevent bottlenecks, they may undermine the effectiveness of entire missions. Precisely, the assumption that prediction models only require a subset of information for image reconstruction may lead to false confidence in a codec to reliably discern the salient signals. We argue the opposite holds, i.e., when the objective is to accommodate \emph{arbitrary} downstream tasks with prediction models instead of human experts, there is \emph{less} potential for rate reductions. Intuitively, two seemingly visually identical images may have subtle differences in pixel intensities, which a prediction model could leverage to overcome physiological restrictions. 

In summary, three conflicting objectives aggravate the challenges for OEC: (i)~maximizing downlinking captures, (ii)~ensuring the value of the captures by relying on as few assumptions on downstream tasks as possible, and (iii)~minimizing the risk from unpredictable adverse effects on current and future prediction models. To this end, we propose drawing from recent work on neural feature compression with Shallow Variational Bottleneck Injection (SVBI)~\cite{es1, es2, frankensplit}. The idea of SVBI is to reduce discarding information necessary for arbitrary, practically relevant tasks by targeting the shallow representation of foundational models as a reconstruction target in the rate-distortion objective. 
%
In other words, rate reductions come from constraining the solution space with abstract high-level criteria rather than reifying target tasks with an explicit definition of value or expert-crafted labels.
We investigate whether the SVBI framework is suitable for EO from a compression perspective and identify lower-level system considerations given the oppressive constraints of OEC. Then, we apply our insights to introduce a \textbf{\underline{T}}ile \textbf{\underline{H}}olistic \textbf{\underline{E}}fficient \textbf{\underline{F}}eatured \textbf{\underline{O}}riented \textbf{\underline{O}}rbital \textbf{\underline{L}}earned (THE FOOL) compression method, which we will refer to as FOOL for short. FOOL alleviates the challenges of OEC by generalizing SVBI to improve compression performance while introducing more specific methods that aid in meeting the requirements of OEC and EO tasks. 
%
%
FOOL comprises a profiler, a neural feature codec with a separate reconstruction model, and a simple pipeline. The profiler identifies configurations that maximize data size reduction, factoring in intermittently available downlinks and the trade-off between processing throughput and lowering bitrate from more powerful but costlier transforms. The neural codec's architecture includes task-agnostic context and synergizes with the profiler's objective to maximize throughput with batch parallelization by exploiting inter-tile spatial dependencies. The pipeline minimizes overhead by CPU-bound pre- and post-processing with concurrent task execution.  
%
%

We perform in-depth experiments to scrutinize our approach with a wide range of evaluation measures by emulating conditions on a testbed with several edge devices. Our results show that FOOL is viable on CubeSat nanosatellites and increases the downlinkable data volume by two orders of magnitude relative to bent pipes at no loss on performance for EO. Unlike a typical task-oriented compression method, it does not rely on prior information on the tasks.
%
Additionally, FOOL exceeds existing SVBI methods with an up to 2.1$\times$ bitrate reduction. Lastly, the reconstruction model can map features from the compressed shallow feature space to the human interpretable input space. The resulting images compete with state-of-the-art learned image compression (LIC) models using mid-to-high quality configurations on PSNR, MS-SSIM, and LPIPS~\cite{lpips} with up to 77$\%$ lower bitrates. We open-source the core compression algorithm\footnote{\url{https://github.com/rezafuru/the-fool}} as an addition to the community. In summary, our main contributions are:
\begin{itemize}
\item Demonstrating the inadequacy of image codecs for EO with satellite imagery and that the general SVBI framework~\cite{frankensplit} can address these limitations.
\item Significantly improving compression performance of existing methods with novel components that embed additional modality and capture inter-tile dependencies of partitioned images. 
\item Introducing a reconstruction component that can recover high-quality human interpretable images from the compressed latent space of shallow features. 
\end{itemize}
To the best of our knowledge, this work is the first to elaborate on the risk of image codecs on EO that distinctly rely on fine-grained details.
Crucially, it proposes a solution approach that assumes reconstruction for human interpretability as a subset of objectives that prioritize maintaining the integrity of model predictions.

\Cref{sec:relwork} compares relevant work addressing the downlink bottleneck. \Cref{sec:motivation} motivates our approach by describing current challenges. \Cref{sec:sysdesign} describes the profiling strategy and compression pipelines. \Cref{sec:method} introduces the FOOL's compression method. \Cref{sec:eval} details the methodology and evaluates FOOL against numerous baselines. \Cref{sec:disclim} transparently discusses limitations to shape directions for future research. Lastly, \Cref{sec:conclusion} concludes the work. 

%% file: sections/related_work.tex
\section{Related Work} \label{sec:relwork}
\subsection{Collaborative Inference and Data Compression} \label{subsec:colinference}
The Deep Learning aspect of our method draws from recent advancements in collaborative inference~\cite{scsurvey} and data compression~\cite{intronc}. The underlying compression algorithm and objective functions are derived and extended from our previous work~\cite{frankensplit}, which re-formulizes the distortion term from lossy compression methods~\cite{ntc} and deploys lightweight models suitable for resource-constrained mobile devices. Besides introducing novel components to further lower transfer costs, FOOL considers the diverging requirements due to intrinsic differences between terrestrial and orbital remote sensing. 
\subsection{Preventing Link Saturation with Orbital Inference} \label{subsec:reloec}
The system aspect of our method aligns best with work focusing on getting the data to the ground for further processing instead of performing inference on board~\cite{giuffrida2020cloudscout, Giuffrida2022phi, denby2020,zhang2024resource}. 
%
We emphasize the high variability among fundamental design principles for OEC, as it is an emerging field, and a comprehensive literature review is not within the scope of this work. 
In summary, we found that current approaches focus on designing complex systems tailored to specific conditions and rely on strong assumptions limiting their applicability. Moreover, they may adequately model the system conditions but run experiments on toy tasks or on low-resolution images. Contrastingly, FOOL is a holistic approach to the downlink problem that considers satellite systems and imagery properties. 
The following discusses the approaches we find most promising as representatives in their general direction. 
%

Gadre et al. introduce \textit{Vista}~\cite{gadre2022low}, a Joint Source Channel Coding (JSCC) system for LoRa-enabled CubeSats designed to enhance low-latency downlink communication of satellite imagery and DNN inference. It shows significant improvements in image quality and classification performance through LoRa-channel-aware image encoding.
Moreover, the evaluation assumes simple tasks that are not representative of practical EO. 
In contrast, FOOL decouples image recovery from the initial compression objective and ensures task-agnostic preservation of information.

Lu et al. introduce \textit{STCOD}~\cite{lu2023satellite}, a JSCC system for efficient data transmission and object detection in optical remote sensing. STCOD integrates satellite computing to process images in space, distinguishing between regions of interest (ROIs) and backgrounds. It shows promising results with a block-based adaptive sampling method, prioritizing transmitting valuable image blocks using fountain code~\cite{fountaincodes}. The caveat is that ROI detectors that can reliably prevent predictive loss for downstream tasks require strong biases regarding sensor and task properties. FOOL includes task-agnostic context, with significantly less overhead and more robustness towards varying conditions than an ROI detector. Furthermore, it is end-to-end optimized with the other compression model components without relying on the same biases or expert-crafted labels. 

Thematically, our work resembles Kodan by Denby et al.~\cite{kodan} the closest. Like FOOL, Kodan treats channel conditions as an orthogonal problem and primarily focuses on source coding to address the downlink and computational bottlenecks. 
Kodan uses a reference application for satellite data analysis and a representative dataset to create specialized small models. Once in orbit, it dynamically selects the best models for each data sample to maximize the value of data transmitted within computational limitations. Kodan's excellent system design is promising but relies on assumptions that hinder practicability and the potential for meaningful rate reductions.
Unlike Kodan, we follow a different design philosophy by treating the downlink bottleneck primarily as a compression problem. Further, we do not treat the computational deadline as a hard temporal constraint to decouple the method to a particular system design, as reflected by FOOL’s profiler measuring key performance indicators on the \emph{pixel level}. Given hardware limitations, the aim is to reduce transfer costs by balancing the lower bitrate of more powerful encoders and the gain in processing throughput of more lightweight encoders. 

%% file: sections/motivation_background.tex
\section{Background \& Problem Formulation} \label{sec:motivation}
\subsection{The Downlink Bottleneck} \label{subsec:oec}
Downlink bottlenecks occur when the data volume exceeds the bandwidth within a downlink segment during a single pass.  
%
We formalize a model sufficient for our purposes by considering link conditions and sensor properties of satellites belonging to a constellation. 
A constellation is defined as $\mathcal{C} = (L, \mathcal{S}, I, f)$ where $L$ is a link to communicate with a ground station and $ \mathcal{S}$ is a set of satellites.  The link is determined by its \textit{expected} downlink rate, measured in megabits per second (Mbps). The function $f: S \rightarrow I$ maps each satellite $s \in S$ to an interval where it passes the downlink segment into disjoint subsets $\mathcal{G} = \{ \mathcal{G}_{i} | \mathcal{G}_{i} = \{s \in \mathcal{S} | f(s) = i \}, i \in I\}$, such that $\bigcup \mathcal{G}_{i} = \mathcal{S}$ and $\bigcap \mathcal{G}_{i} = \varnothing$. The link capacity $V_{\text{link}}$ is the bandwidth available per pass and is determined by the link rate and the interval range.
Satellites $S=(R_{\text{orbit}}, S_{\text{rate}}, S_{\text{spatial}}, S_{\text{bands}}, S_{\text{radio}}, S_{\text{fov}})$ are equipped with a sensor, and its properties determine the volume per capture. 
\input{embed/equations/volume_captures}
The radiometric resolution $S_{\text{radio}}$ and number of bands $S_{\text{bands}}$ determine the downlink cost per pixel in bits. The orbit $R_{\text{orbit}}$, sensor spatial resolution $S_{\text{spatial}} = S_h \times S_w$, and field of view $S_{\text{fov}}$ determine the number of pixels per capture. The number of captures depends on the time to complete an orbit
\input{embed/equations/time_to_orbit}
and on the capture rate $S_{\text{rate}}$. $G$ is the gravitational constant and $M$ is the earth’s mass. The orbit $R_{\text{orbit}}$ is usually around 160 to 800 kilometers for LEO satellites. For reference, $R_\text{orbit}$ is 786 kilometers for Sentinel-2~\cite{sentinelsensors}.
Finally, the number of captures from all satellites within the segmentation group determines the total volume per pass.
\input{embed/equations/volume_pass}
The superscript $(i)$ denotes the costs associated with a satellite $(i)$. For constellations with homogenous sensors $V_{\text{capture}}$ is a static value. Notice that $V_\text{pass}$ scales linearly by the constellation size and a constant factor $c$ for overlap occurrences, i.e., $|\mathcal{G}_{j}| = \frac{|\mathcal{S}|}{c}$. To determine $c$ for a constellation, we must calculate the minimum angle between satellites $\beta^{*}$. Assuming a single ground station at the Earth's North Pole and given the minimum communication elevation $\theta$
\input{embed/equations/min_angle_satellites}
For example, consider a constellation at $R_{\text{orbit}} = 790,000$ meters altitude with a minimum elevation 
$\theta = 25^{\circ}$
such that $\beta^{*} \approx 22.52^{\circ}$. Then, $c = \frac{360^{\circ}}{22.52^{\circ}} \approx 16$, i.e., to prevent any interval sharing, the constellation size may not exceed 16 satellites.

In short, the aim is to facilitate cost-efficient scaling of constellations by increasing bandwidth value and substantially reducing reliance on building additional infrastructure. That is, we require an encoding scheme $\texttt{enc}$, such that $V_{\texttt{enc}} < V_{\text{link}}$.
Note that a single satellite may experience a bottleneck even if the constellation is sparse enough to prevent interval sharing~\cite{kodan}. Say, each $s \in \mathcal{S}$ is equipped with a sensor using approximate Sentinel-2 configurations~\cite{sentinelsensors} by setting a multispectral sensor for (near-) visible light to $S_{\text{bands}} = 4$, $S_{\text{radio}} = 12$ $S_{\text{fov}} = 21^{\circ}$, $S_{h\times w} = 10 \times 10$, and five captures per pass. With $|\mathcal{S}| \leq 16$, the volume for each pass is $\frac{790,000^2 \cdot \operatorname{tan}^2(10.5^{\circ})}{100} \cdot 4 \cdot 12 \cdot 5 \approx 410$ GB. To prevent a bottleneck even without sharing an interval and using a higher-end link, such as WorldView-3~\cite{worldview3} where $V_{\text{link}} = 90$ GB per pass, the $\texttt{enc}$ needs to decrease the data volume by a factor of 4.5. 

There are two overarching objectives for a codec and the system we deploy its encoder. The system’s objective is to process and encode large volumes of high-dimensional data, given the physical limitations of LEO (nano-) satellites. A 3U nanosatellite following the CubeSat standard is limited to 10cm$\times$10cm$\times$30cm and 4kg~\cite{mehrparvar2022cubesat} with restricted power supply by using solar harvesting~\cite{denby2019orbital}. The compression objective is to achieve a sufficiently low bitrate while maintaining the data’s integrity. The following elaborates on the challenges of conceiving a method that fulfills our criteria and the limitations of applying existing codecs.
\subsection{Limitations of Codecs} \label{subsec:limexistingcodecs}
Given remote image captures and a set of unknown associated object detection tasks,
we seek a transformation of the captures into representations that minimize transfer costs and loss of information that may impact any detection tasks.
We refer to generalizability as a measure of how well a method can minimize the predictive loss on unknown detection tasks. For example, a purely task-oriented encoding (e.g., \cite{singhfc}) can retain information for a set of explicitly defined tasks. Still, it does not generalize as the transformed data is unusable for non-overlapping tasks
Besides bent pipes, \textit{lossless} codecs are the only 
approach with easily understood guarantees on generalization. Nevertheless, lossless compression cannot adequately address the downlink bottleneck due to theoretical lower bounds. Promising alternatives are \textit{lossy} methods that relax the requirement of relying on identical reconstruction for generalization. More formally, given a distortion measure $\mathcal{D}$, a constraint $D_c$ bounds the minimal bitrate to~\cite{shannon1959coding}:
\input{embed/equations/rd_objective}
where $I(X;Y)$ is the mutual information and is defined as:
\input{embed/equations/mutual_inf}
Learned Image Compression (LIC) replaces the typically linear transformation of handcrafted codecs with nonlinear ones to reduce dependencies from sources that are not jointly Gaussian~\cite{ntc}. The sender applies a parametric analysis transform $g_a(\boldsymbol{\mathrm{x}}; \theta)$ into a latent $\boldsymbol{\mathrm{y}}$, which is quantized to a latent with discrete values $\hat{\boldsymbol{\mathrm{y}}}$. Then, an entropy coder losslessly compresses $\hat{\boldsymbol{\mathrm{y}}}$ using a shared entropy model $p_{\hat{\boldsymbol{\mathrm{y}}}}$. The receiver decompresses $\hat{\boldsymbol{\mathrm{y}}}$ and passes it to a parametric synthesis transform $g_s(\hat{\boldsymbol{\mathrm{y}}};\phi)$ to recover a distorted approximation to the input. To capture leftover spatial dependencies of $\hat{\boldsymbol{\mathrm{y}}}$, more recent work adds \emph{side information} with a hyperprior $\boldsymbol{\mathrm{z}}$~\cite{shp} and a context model~\cite{jahp}. Including side information requires two additional parametric transforms $h_a(\hat{\boldsymbol{\mathrm{y}}};\theta_h)$ and $h_s(\hat{\boldsymbol{\mathrm{z}}};\psi_h)$. 
Despite efficient LIC methods~\cite{tinylic} consistently outperforming handcrafted codecs on standardized benchmarks, the results are deceptive when assessing the impact on downstream tasks. To provide further explanation, we perform a preliminary experiment that contrasts the predictive loss with additive noise and codec distortion and summarize results in \Cref{plot:lossyvsnoisynoft}.  

\input{embed/media/plot_noisy_vs_lossy}
We download pre-trained weights~\cite{wightman2021resnet} for the ImageNet~\cite{imagenet} classification task of three popular architectures\cite{he2015deep, swin2021, liu2022convnet}.
First, we compute the expected Peak Signal-To-Noise Ratio (PSNR) of a popular LIC model~\cite{fp} for each quality level on the validation set. Then, we apply Additive White Gaussian Noise (AWGN) on input to match the PSNR of a codec for each quality level separately. Lastly, we measure the \textit{predictive loss} as the average difference between the accuracy of the original and processed samples. 
Notice that the predictive loss on the distorted input is significantly worse than the noisy input. 
Additive noise does not remove information; rather, it superimposes unwanted information. Conversely, lossy compression intentionally discards information from signals, and two codecs may achieve comparable rate-distortion performance despite emphasizing different information to retain. Re-training model weights on reconstructed samples may mitigate some predictive loss, but only due to adjusting prediction to input perturbations and error-prone extrapolation of lost information.

Particularly, for EO with satellite imagery that spans large geographic areas, we stress the unsuspecting danger of lossy compression, which is compounded with learned transforms~\cite{ntc}, where it is challenging to understand behavior.
%
The ability to differentiate between intensities beyond the capability of humans may explain why detection models can outperform domain experts. Accordingly, we should assume that lossy codecs may discard information where even experts cannot reliably verify the impact on machine interpretability. For example, suppose a codec that reduces the rate by focusing on preserving coarser-grained structures. Then, tasks that rely on assessing the environment for fine-grained object classes will lack background information (e.g., inferring region by tree species with subtle color variations).

\input{embed/media/fig_hd}
Current limitations of image codecs put operators in a difficult position, especially for EO. The decision falls between (a combination of) lossless codecs, applying crude filtering methods, or attempting to reduce the bitrate with lossy codecs, remaining uncertain about whether the codecs retain information necessary in real conditions. 
As a solution, we advocate for \textit{Shallow Variational Bottleneck Injection} (SVBI)~\cite{frankensplit}, which prioritizes salient regions for (near) arbitrary high-level vision tasks.
\subsection{Shallow Variational Bottleneck Injection} \label{subsec:vib}
SVBI trains neural codecs by replacing the distortion term of the rate-distortion objective of variational image compression models~\cite{shp} with head distillation (HD)~\cite{hd1, hd2, es1, es2, frankensplit}.  \Cref{fig:hd} illustrates an example where the HD distortion measure penalizes a compression model for not sufficiently approximating the shallow representation of a pre-trained foundational model. We define a foundational model as a pre-trained DNN that can accommodate multiple tasks by attaching predictors or fine-tuning the deeper layers. 
In Knowledge Distillation (KD) terminology, the codec is referred to as the \emph{student} and the shallow layers of a foundational model as the \emph{teacher}. Note that KD is not this work's focus, as HD diverges from the typical KD objective. 
The intuition behind SVBI is that if a codec can reconstruct the representation of a foundational model, then the representation is sufficient for \emph{at least} all tasks associated with that model. 
\subsubsection{The Effectiveness of Shallow Features} \label{subsubsec:genshallowfeatures}
Readers may reasonably assume that using the representation for one particular network architecture instead of the input as the distortion measure is more restrictive for two reasons.

\input{embed/media/fig_recon_motivational}
First, the features are not human-readable, i.e., we cannot overlay the bounding boxes on the images. We could infer the global coordinates to present boxes overlaid on previously captured satellite imagery. This may suffice for observing (semi-) permanent objects (e.g., landmarks) but certainly not for ephemeral or moving objects (e.g., tracking the movement of vessels).
Second, even if the trend toward transfer learning with foundational models~\cite{awais2023foundational, tlearningfoundational} can accommodate various predictors, client preferences for architectures may vary. 

We argue that by targeting shallow representations, both limitations can be addressed. 
View an $n$-layered feed-forward neural network as a Markov chain of successive representations $R_{i}, R_{i+1}$~\cite{tishbyinfomc}:
\input{embed/equations/markovrepr}
The mutual information $I(X;R_i)$ will likely decrease relative to the distance between the input and a representation. This loss stems from layers applying operations that progressively restrict the solution space for a prediction, particularly for discriminative tasks. That is, the deeper the representation, the more information we lose regarding  $X$: 
\input{embed/equations/sampleinfoloss}
\Cref{fig:reconmotivational} visualizes the trade-off. We extract the features from a ResNet network with weights trained on the ImageNet~\cite{imagenet} classification task and recover the original image by training separate reconstruction models for each marked location. \Cref{subsec:imagerecon} elaborates on the reconstruction. Notice that models at shallow layers can recover the input with high similarity, but the recovery progressively worsens as the path distance increases. Now assume a discriminative model (e.g., a ResNet for image classification) $\mathcal{M} = (\mathcal{H}, \mathcal{T})$ with separate shallow $\mathcal{H}$ from deeper layers $\mathcal{T}$ as disjoint subsets, such that $\mathcal{H}(X) = H$ (i.e, mapping input to shallow features) and $\mathcal{M}(X) = \mathcal{T}(\mathcal{H}(X))$. Further, given a codec $\mathrm{c} = (\mathrm{enc}, \mathrm{dec})$ where $\mathrm{dec}(\mathrm{enc}(X)) = \tilde{H}$ is an approximation of $\mathcal{H}(X)$. Then, $\tilde{H}$ is a sufficient approximation of $H$ if $\mathcal{T}(\tilde{H})$ results in lossless prediction, i.e., no drop in prediction performance relative to $\mathcal{T}(H)$. In other words, a sufficient representation in the shallow latent space results in high similarity in the deep latent space between $\mathcal{T}(H)$ and $\mathcal{T}(\tilde{H})$. 
%
Consider that similarity in the deep latent space coincides with high similarity for human perception in the input space~\cite{lpips}. 
Therefore, the encoder output $\texttt{enc}(X)$ should retain sufficient information to reconstruct $X$ with quality comparable to $\mathcal{H}(X)$, as exemplified in \Cref{fig:reconmotivational}. Finally, since $\texttt{enc}(X)$ sufficiently approximates $H$, it should be possible to sufficiently approximate the shallow representation of any model $\mathcal{M}^{\prime} = (\mathcal{H}^{\prime}, \mathcal{T}^{\prime})$ if $I(\mathcal{H}(X), X) \approx I(\mathcal{H}^{\prime}(X), X)$. 
%
\subsubsection{Rate Reductions by Task Specificity} \label{subsubsec:machineandhumanperception}
%
The idea of task-oriented communication is that messages for model prediction may require less information than human domain experts, i.e., that it should be possible to reduce bitrate by not (exclusively) using input reconstruction as the distortion measure. 
We argue that this assumption contradicts empirical evidence demonstrating models outperforming human experts in various image-related tasks, i.e., machines can detect signals and patterns that humans physiologically or intellectually cannot. Rather, the opposite should hold, i.e., when compressing for quality using domain experts as judges, we should see more potential for rate savings, not less. The claim is consistent with the results in \Cref{plot:lossyvsnoisynoft} where codecs with high reconstruction quality result in images that are deceptively similar to the input (details in \Cref{subsec:compperfhuman}). 
Conclusively, rate reductions are from \emph{task specificity} of the distortion measure, irrespective of the input interface, whether it is a particular layer of a DNN architecture, human receptors, or textual encoding. Note that this holds, even if when limiting measures to discriminative task objectives without any image reconstruction. 
Besides visualizing \Cref{eq:sampleinfoloss}, the input image illustrates a practical example. The frog subset of ImageNet distinguishes between \textit{Tree Frogs}, \textit{Bullfrogs}, and \textit{Tailed Frogs}. Since these frog species have distinct figures and dominant colors, the more delicate characteristics of a tree frog are redundant for ImageNet classification. The network gradually discards information regarding the fine-grained blue-yellow colored patterns, permitting only the recovery of general shape and environment from the deep features. The deeper the features, the less structure and detail are present, which may be redundant for the task. 
%
Now, suppose training a codec where the encoder retains the minimal information necessary to reconstruct the output of the deepest layers for a classification task (e.g., similar to Vista~\cite{gadre2022low}). Then, we can reduce the transfer cost to as low as $\log_2{({\#\text{labels}})}$ without predictive loss. However, we may lack the information for other tasks, i.e., there is a trade-off between generalization and the lower bound on the bitrate. 
In contrast, targeting shallow features for compression may strike a balance between aiming to retain information for all possible downstream tasks and only emphasizing the salient regions for the tasks associated with a foundational model. Arguably, the limitation is negligible, as maintainers will train foundational models with useful tasks in mind.
%

%% file: embed/equations/volume_captures.tex
\begin{equation}
V_{\text{capture}} = \overbrace{\frac{R^{2}_{\text{orbit}} \cdot \operatorname{tan}^2(S_{\text{fov}})}{S_w \cdot S_h}}^{\text{Total Pixels}} \cdot \underbrace{S_{\text{bands}} \cdot S_{\text{radio}}}_{\text{Bits per Pixel}}
\end{equation}

%% file: embed/equations/time_to_orbit.tex
\begin{equation}
T_{\text{orbit}} =2\pi \sqrt{\frac{(R_{\text{orbit}} + R_{\text{earth}})^{3}}{GM}}
\end{equation}

%% file: embed/equations/volume_pass.tex
\begin{equation}
V_{\text{pass}}  = \sum_{s^{(i)} \in \mathcal{G}_{j}} T_{\text{Orbit}} \cdot S^{(i)}_{\text{rate}} \cdot V^{(i)}_{\text{capture}}
\end{equation}

%% file: embed/equations/min_angle_satellites.tex
\begin{equation}
\beta^{*} = 2 \times (180^{\circ}-(\theta + 90^{\circ}) - \arcsin({\frac{R_{\text{earth}} \cdot \sin{(90^{\circ}+\theta)}}{R_{\text{orbit}} + R_{\text{earth}}}})
\end{equation}

%% file: embed/equations/rd_objective.tex
\begin{equation}
\underset{P_{Y|X}}{\mathrm{min}}\; I(X;Y)\; \text{s.t.}\; \mathcal{D}(X,Y) \leq D_c \enspace ,
\label{eq:rdbasics}
\end{equation}

%% file: embed/equations/mutual_inf.tex
\begin{equation}
I(X;Y)=\int_{}\int_{} p(x,y) \log \left(\frac{p(x,y)}{p(x) p(y)}\right) dxdy \enspace .
\label{eq:mutualinf}
\end{equation}

%% file: embed/media/plot_noisy_vs_lossy.tex
\begin{figure}[htb]
    \centering
    \includegraphics[width=\columnwidth]{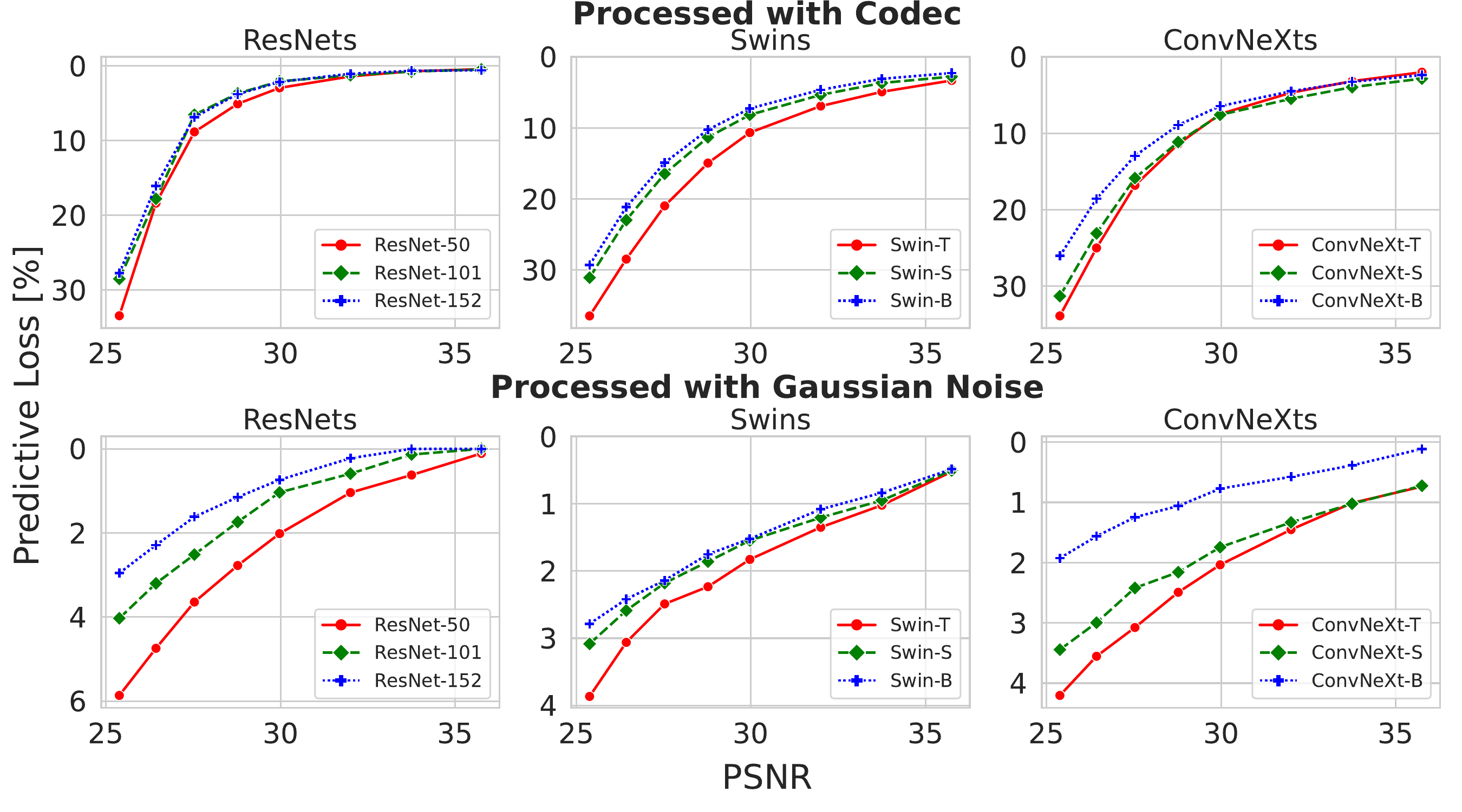}
    \caption{Comparing Effect of Codec and Additive Noise}
    \label{plot:lossyvsnoisynoft}
\end{figure}

%% file: embed/media/fig_hd.tex
\begin{figure}[htb]
    \centering
    \includegraphics[width=\columnwidth]{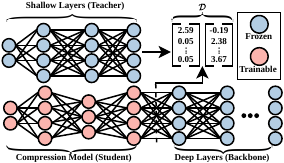}
    \caption{Head Distillation Distortion Loss}
    \label{fig:hd}
\end{figure}

%% file: embed/media/fig_recon_motivational.tex
\begin{figure}[htb]
    \centering
    \includegraphics[width=\columnwidth]{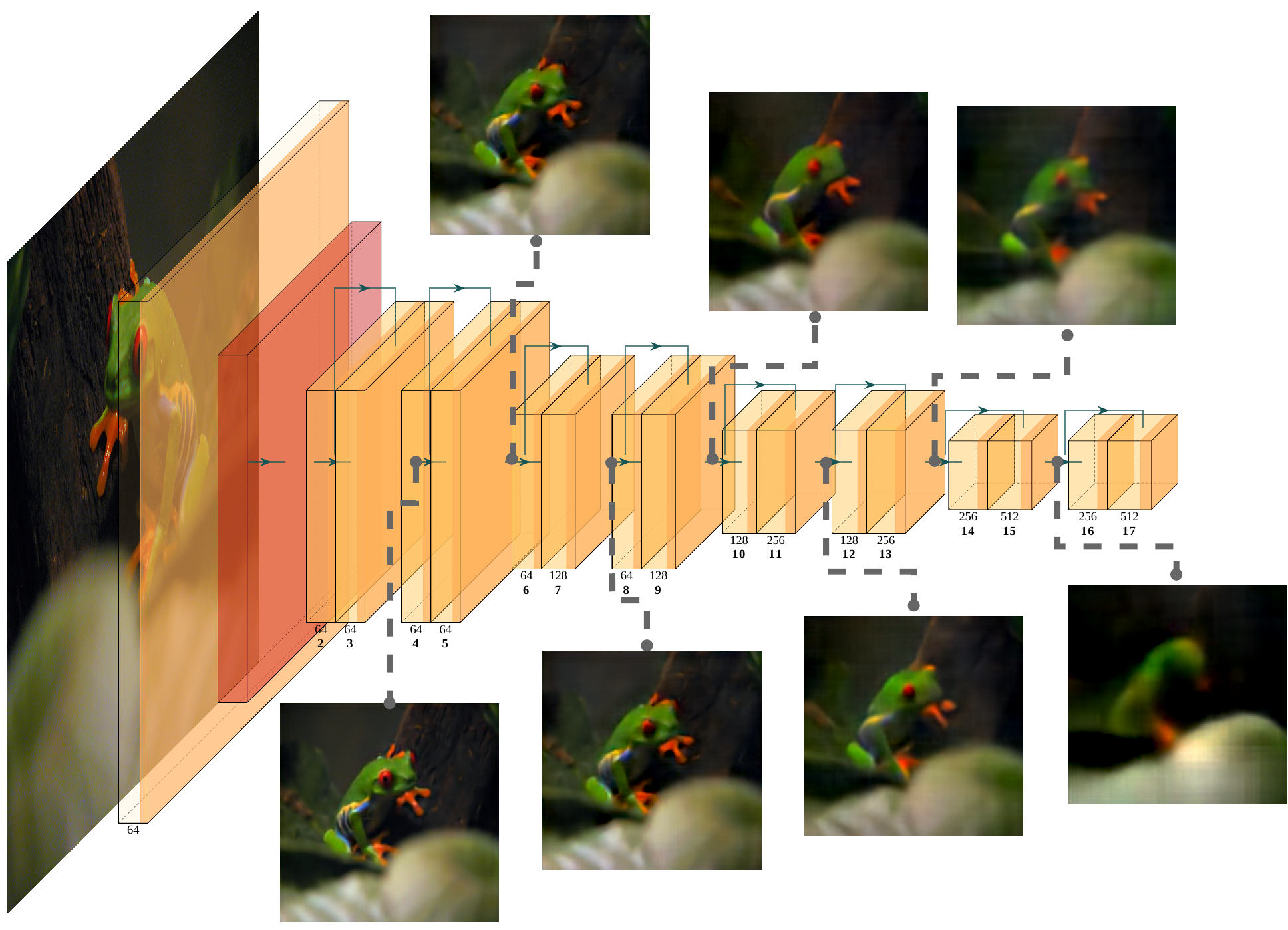}
    \caption{Discarding Information in Discriminative Tasks}
    \label{fig:reconmotivational}
\end{figure}

%% file: embed/equations/markovrepr.tex
\begin{equation}
\label{eq:dnnmarkovchain} 
I(X ; Y) \geq I\left(R_1 ; Y\right) \geq \ldots \geq I\left(R_n ; Y\right) \geq I(\tilde{Y} ; Y)
\end{equation}

%% file: embed/equations/sampleinfoloss.tex
\begin{equation}
\label{eq:sampleinfoloss} 
I(X; X) \geq I(R_{1}; X) \geq \dots \geq I(R_{n-1}; X) \geq I(R_{n}; X) 
\end{equation}

%% file: sections/method_systems.tex
\section{The FOOL’s System Design} \label{sec:sysdesign}
\subsection{Compression and Prediction Request Flow} \label{subsec:reqflow}
\Cref{fig:foolhlvl} illustrates a high-level view for serving requests.

\input{embed/media/fig_fool_request_flow}
For samples processed by FOOL, there is a single encoder. The output $\boldsymbol{\hat{{\mathrm{y}}}}$ 
%
is forwarded to the detection pipeline, skipping the shallow layers. The detection pipeline for a single forward pass consists of a decoder, backbone and predictor. There may be multiple backbones the client can choose from, and each backbone may have multiple predictors. A decoder transforms $\boldsymbol{\hat{{\mathrm{y}}}}$ into an input representation for a particular backbone. A predictor outputs bounding boxes for a specified task. An image reconstruction model optionally restores the latent to a human-interpretable image to overlay the bounding boxes. Samples downlinked with bent pipe or some image codec are forwarded to the shallow layers, skipping the corresponding decoder and reconstruction model. This section focuses on the pipeline, before \Cref{sec:method} introduces the compression method.
\subsection{Profiling Compression Pipelines for OEC} \label{subsec:resourcemaximation}
A common challenge for operators is to determine whether reported performance regarding resource usage or throughput from the latest advancements generalizes to their target hardware. This problem stems not from a lack of rigor by authors but from the sheer heterogeneity of the AI accelerator landscape~\cite{reuther2022ai}. Graph compilers and other vendor-specific optimizations (e.g., TensorRT\footnote{https://developer.nvidia.com/tensorrt}, Apache TVM\footnote{https://tvm.apache.org}) further complicate evaluation, with varying methods for operator fusion, graph rewriting, etc. Consequently, FOOL includes a simple profiling and evaluation strategy that operators may run before deployment.
Notably, in contrast to existing work that partitions images to match the input size of a particular application, the profiler regards the importance of spatial dimensions for resource efficiency. The purpose of the profiler is to determine a configuration that maximizes throughput. While throughput evaluation is straightforward, how to measure it (e.g., images/second) is not necessarily obvious, particularly for (neural) compression pipelines. 

First, terrestrial and LEO remote sensing with constrained sensor networks demand resource-conscious methods, but in LEO, downlinks are only available within segments. Due to memory and storage constraints, devices must process samples according to a sensor rate, i.e., a prolonged interval between incoming samples. Hence, the objective in LEO is to maximize the number of pixels the accelerator can process before reaching a downlink segment, given a time constraint for a single sample (i.e., ``frame deadline''~\cite{kodan}). For example, assume a cheaper and a costlier compression model where both models meet the frame deadline. Applying the latter results in half the bitrate but thrice the inference time. Using the former is beneficial in most network conditions for real-time terrestrial applications since it results in a lower end-to-end request latency. In contrast, applying the latter in LEO may be advantageous, as finishing earlier results in the needless idle time of resources. 
Second, satellite imagery has substantially higher resolution than captures from most terrestrial sensor networks. A standard method to improve throughput for high-dimensional images is parallel processing with tile partitioning. The distinction is that there is more control over the spatial dimensions and the batch size. Nonetheless, a caveat is the friction between a model’s size and the input size. Increasing the width (e.g., the number of feature maps output by a convolutional layer) of a neural codec's parametric transforms may result in better compression performance but lower processing throughput. In summary, we require a measure that includes (i) the tile spatial dimensions, (ii) batch size, and (iii) the capacity-compression performance trade-off. 

We can address the requirements (i) and (ii) by measuring throughput as \textit{pixels processed per second} (PP/s). To motivate the need to expand on PP/s for (iii), we demonstrate the friction between model width, input size, and batch size using the convolutional encoder in \cite{frankensplit} consisting of three downsampling residual blocks (\Cref{subsec:contextforfcomp}). \Cref{plot:throughputpilot} summarizes the results as the average of 100 repetitions with progressively increasing width.
\input{embed/media/plot_throughput_pilot}
Notice how evaluating img/s always favors smaller spatial dimensions and disregards batch size and model width. In contrast, PP/s reveals that the optimal spatial dimension is around $500 \times 500$ but will naturally favor smaller models, as it does not consider that wider models may reduce transfer costs. To alleviate the limitations of PP/s, we measure \textit{Transfer Cost Reduction per Second} (TCR/s) as:
\input{embed/equations/transfer_cost_reduced_per_seconds_sub}
The measure now includes the compression performance as the 
difference 
between the expected bits per pixel (bpp) of compressed ($\text{bpp}_{\text{enc}}$) and uncompressed ($\text{bpp}_{\text{raw}}$) sensings. 
The raw bpp value refers to the bit depth, i.e., the sensor’s radiometric resolution and the number of bands. For example, the radiometric resolution of Sentinel-2 is 12 bits~\cite{sentinelsensors}, so for three bands $\text{bpp}_{\text{raw}}=3 \cdot 2^{12}$. 
%
The advantage of TCR/s is twofold. First, decoupling it from system-specific parameters, such as sensor resolution or orbital period, permits drawing generalizable insights regarding the relative trade-off between codec overhead and bitrate reduction. Second, operators can still assess the feasibility of the pipeline on target hardware and the expected downlinkable data volume by running the profiler with configurations that reflect deployment conditions.
\subsection{Concurrent Task Execution} \label{sec:conctask}
So far, this section has solely discussed the computational cost of a codec's parametric transforms without considering pre- and post-processing. In particular, after applying the encoder transforms, it is still necessary to entropy code the output to compress the latent. Since FOOL's entropy model is input adaptive, it requires a range coder. Although more recent range coders are efficient, they incur non-negligible runtime overhead. Therefore, given the unforgiving conditions of OEC, we argue that the entropy coder cannot be neglected in the design process and evaluation of a neural codec. FOOL virtually offsets the entire runtime overhead with simple concurrent task execution. The idea is to exploit the minimal interference of processes that draw from different resource types.
For three sequentially incoming samples $x_{i-1}, x_{i}, x_{i+1}$, FOOL executes CPU-bound pre-processing of $x_{i+1}$, accelerator-bound inference of $x_{i}$, and CPU-bound post-processing of $x_{i-1}$.  In this work, pre-processing corresponds to tiling the samples, and post-processing to entropy coding with rANS~\cite{ans, rans}. 
Concurrently to inference ${x_{i-1}}$ on the accelerator, a process starts tiling $x_i$. After inference on ${x_{i-1}}$, $\boldsymbol{\hat{\mathrm{y}}},\boldsymbol{\hat{\mathrm{z}}}, \boldsymbol{\hat{\sigma}}, \boldsymbol{\hat{\mu}}$ (\Cref{subsubsec:trainingobj}) are persisted on the file system. Then, a separate process loads the data and losslessly compresses $\boldsymbol{\hat{\mathrm{y}}},\boldsymbol{\hat{\mathrm{z}}}$ with an entropy coder.
%
We expect minimal interference between the processes, resulting in virtually no PP/s decrease. 

%% file: embed/media/fig_fool_request_flow.tex
\begin{figure}[htb]
    \centering
    \includegraphics[width=\columnwidth]{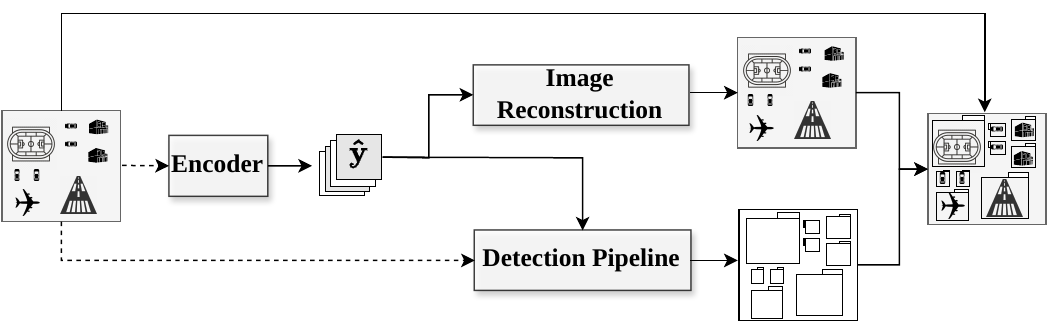}
    \caption{High-Level Inference Request Flow}
    \label{fig:foolhlvl}
\end{figure}

%% file: embed/media/plot_throughput_pilot.tex
\begin{figure}[htb]
    \centering
    \includegraphics[width=\columnwidth]{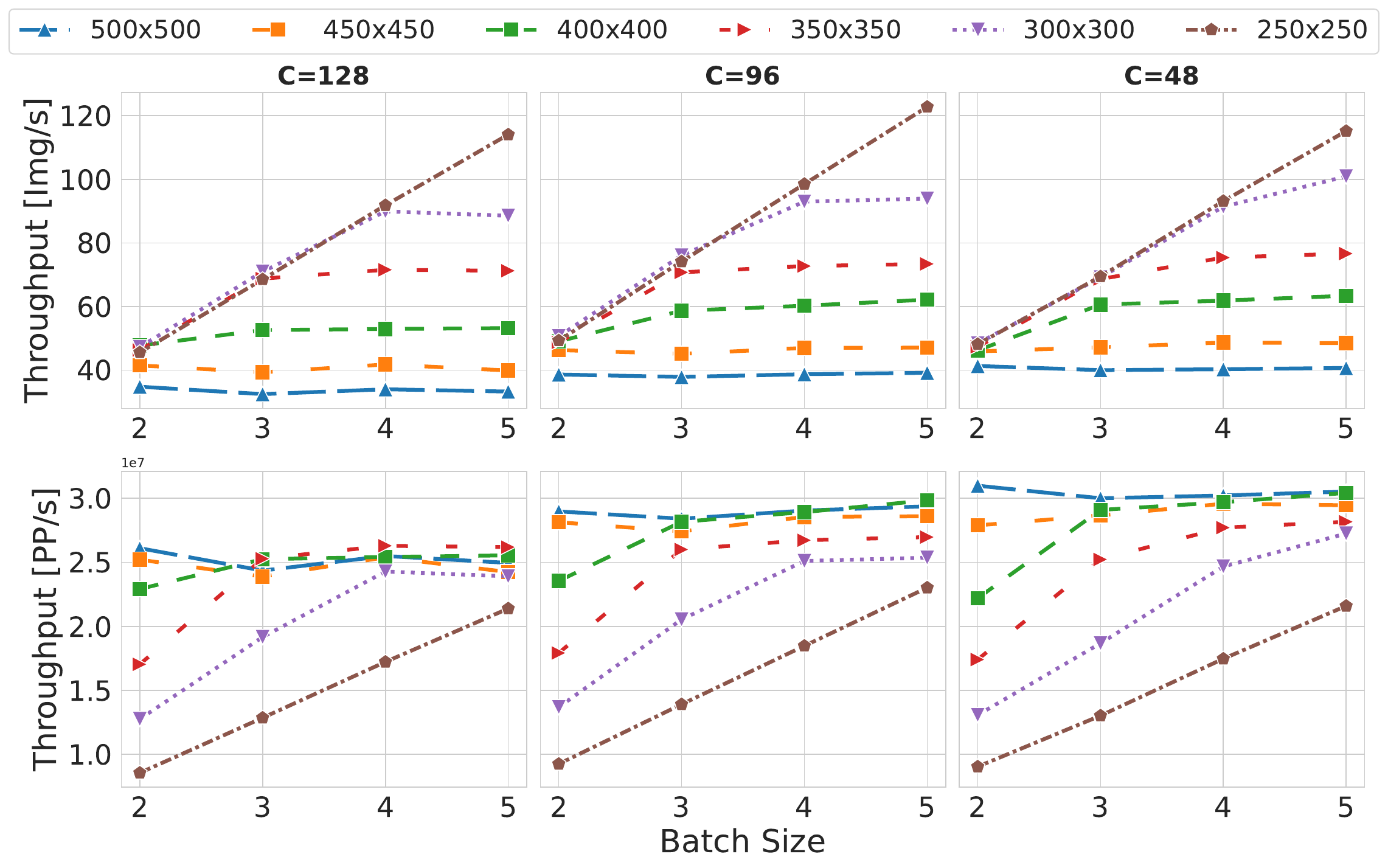}
    \caption{Contrasting Throughput Measures}
    \label{plot:throughputpilot}
\end{figure}

%% file: embed/equations/transfer_cost_reduced_per_seconds_sub.tex
\begin{equation}
\text{TCR/s} = \underbrace{\frac{\text{Image Dimension}}{\text{Seconds per Batch}}}_{\text{PP/s}} 
\times (\text{bpp}_{\text{raw}} - \text{bpp}_{\text{codec}})\label{eq:tcrs}  
\end{equation}

%% file: sections/method_compression.tex
\section{The FOOL's Compression Method} \label{sec:method} 
We design the compression method based on three criteria. First, it should synergize with the profiling strategy (\Cref{subsec:resourcemaximation}). Second, it should embed context for feature compression without favoring a particular downstream task. Third, it should prioritize the integrity of downstream tasks but allow recovering human interpretable images without increasing the bitrate.
\subsection{Model Building Blocks} \label{subsec:modelbuildingblocks}
For a focused evaluation and transparent discussion on the efficacy of our contributions in \Cref{sec:eval}, we restrict FOOL to basic layer types and exclude methods from work on efficient neural network design (e.g., dilated convolutions to increase the receptive field). 
Moreover, basic layer types ensure widespread support across hardware vendors~\cite{reuther2022ai}.  
\input{embed/media/fig_arch_components_org}

\Cref{fig:archorgcomp} illustrates the building blocks of the codec architecture we will introduce in \Cref{subsec:comparch} and how it organizes the primary networks for transform coding (\Cref{subsec:limexistingcodecs}). The primary networks have four stages that control the depth and width. Each stage has at most one lightweight attention block and at least one residual block. A residual block optionally up or downsamples the spatial dimensions. A stage's width and depth parameters configure the number of channels and residual blocks within a stage. 
\subsection{Capturing Inter-Tile Dependencies} \label{subsubsec:tilecorrelations}
The input to the compression model are tiles that were partitioned to maximize processing throughput (\Cref{subsec:resourcemaximation}), i.e., we consider an input $\boldsymbol{\mathrm{x}}$ as a list with $T$ separate image tensors $x_t \in \mathbb{R}^{C \times H \times W}$. 
To decrease transfer costs further, FOOL leverages the prior knowledge from partitioning (i.e., tiles corresponding to the same image) in two ways.
The first is via weight-sharing with 2D Residual Blocks by reshaping the tensor to $T \cdot B \times C \times H \times W$. This way, we include further inductive bias during training by forward passing $T$ similar tensors before each backpropagation. 
The second is with an inter-tile attention mechanism. Since self-attention from transformer architectures is prohibitively expensive for our purposes, even when applying it on downsampled representations as proposed in~\cite{vqgan}.
Therefore, we modify and extend the lightweight convolutional attention layer from \cite{chenglearned}. The layer stacks residual units to increase the receptive field that primarily emphasizes local interactions.
%
FOOL partially replaces the need for global operations by assuming tiles to have ``pseudo-temporal’’ dependencies. Intuitively, partitioning single captures that span large geographic areas may be similar to moving a video feed with large strides. In particular, tiles within the same regions or biomes have global dependencies.
In the 3D version of the attention block from \Cref{fig:archorgcomp}, a residual unit consists of two $1 \times 1 \times 1$ convolution and a $D \times 3 \times 3$ convolution in between. The kernel size for the temporal dimension of attention layers (\Cref{subsec:modelbuildingblocks}) is set as $D=3$ for $T < 5$ and $D=5$ for $T \geq 5$.

The advantage of 3D layers over concatenating channels and applying 2D convolutional operations (e.g., with channel attention~\cite{guo2022attention}) is that it considerably reduces width. For example, given a $3 \times 3$ 2D convolution with $T \times C$ in and out channels. Then for $T=5$ image tensors, with dimensions $C=64, H=128, W=128$, would require $5 \cdot 64 \cdot (3 \cdot 3 \cdot 5 \cdot 64 + 1)= 921,920$ parameters. Conversely, for a $5 \times 3 \times 3$ 3D convolution with the same in and out channels, it would result in $64 \cdot (5 \cdot 3 \cdot 3 \cdot 64 + 1) = 184,384$. Additionally, we reduce the number of multiply-and-accumulates from approximately 15 million to 9 million. 
Besides lowering memory requirements, this allows FOOL to scale model capacity with less friction against processing throughput.
\subsection{Task-Agnostic Context for Feature Compression} \label{subsec:contextforfcomp}
The leftover spatial dependencies after encoding are commonly around high-contrast areas. 
Consider that high-contrast areas typically correspond to edges and other regions of interest, i.e., \textit{keypoints}. 
\input{embed/media/fig_leftoverdep_vs_keypoints-whmap}

As an example, \Cref{fig:leftovervskeypoints} contrasts leftover pixel dependencies of $\hat{y}$ from a LIC model~\cite{jahp} to keypoints output by a KeyNet~\cite{keynet} network. Therefore, we should further improve compression performance with side information by embedding keypoints as context for encoding as follows:
\input{embed/equations/sideinfo_with_kps}
where $k$ is a keypoint extraction function $k: \mathbb{R}^{3 \times H \times W} \rightarrow \mathbb{R}^{1 \times H \times W}$ , $f_{ds}$ is a parametric downsampling function $f_{ds}: \mathbb{R}^{1 \times H \times W} \rightarrow \mathbb{R}^{C^{\prime} \times \frac{H}{2^{n}} \times \frac{W}{2^{n}}}$, and $a_c$ is a single (2D) cross-attention block (\Cref{fig:archorgcomp}). 
The cross-attention block takes context as an additional input for weighting the latent with attention scores. 
For $k$, we use scores from a (frozen) pre-trained and simplified KeyNet~\cite{keynet} due to its robustness in diverse environments and low memory requirements (less than $6000$ parameters). 
While this method should generalize to LIC, it complements feature compression exceptionally well and found 
quantizing and compressing $\boldsymbol{\mathrm{y}}_{ca}$ (i.e., $\boldsymbol{\tilde{\mathrm{h}}}=g_s(Q(\boldsymbol{\mathrm{y}}_{ca}),\psi)$ further lowers the bitrate without affecting  task performance. 
%
\subsection{Compression Model Architecture} \label{subsec:comparch}
\Cref{fig:foolcomparch} illustrates the compression model’s complete architecture. The dashed lines to $Q$ indicate that we either quantize (and subsequently compress) the base latent or the cross-attention weighted latent. 
\input{embed/media/fig_fool_comp_arch}

For the case of passing $\boldsymbol{\mathrm{y}}_{ca}$ to $Q$, we include an additional residual unit after attention-weighting.
We skip applying the attention block to the highest input dimensions to reduce memory and computational costs. 
The non-linearity between the layers is ReLU to reduce vendor dependency of results from the system performance evaluation in \Cref{subsec:sysperf}.  
Residual blocks are two stacked $3 \times 3$ convolutions to increase the receptive field with fewer parameters and a residual connection for better gradient flow.
For the remainder of the work, we refer to the compression model as an encoder-decoder pair $enc, dec$. The encoder comprises $g_a, h_a, k, f_{ds}, c_a$, and the entropy coder. The decoder consists of the $g_s$ and the entropy decoder. The entropy model $p_{\boldsymbol{\hat{\mathrm{z}}}}(\boldsymbol{\hat{\mathrm{z}}})$ and $h_s$ are shared. Note that, despite deploying more components on the constrained sender, the encoder has significantly fewer parameters than the decoder since we increase the width of the receiver-exclusive components.
%
\subsection{Single Encoder with Multiple Backbones and Tasks} \label{subsec:encmultisupport}
Analogous to \cite{frankensplit}, consider a set of $n$ shallow and deep layers pairs of backbones (i.e., foundational models):
\input{embed/equations/bbpairs}
The shallow layers map a sample to a shallow representation, i.e., $\mathcal{H}_i(x) = h_i$. Further, associate a separate set of $m$ predictors $\mathcal{P} = {\mathcal{P}_1, \dots \mathcal{P}_m}$ to the non-shallow (i.e., deep) layers of a backbone. 
Assume an encoder-decoder pair can sufficiently approximate the shallow layer’s representation of a particular backbone (i.e., $dec(enc(\mathrm{\boldsymbol{x}})) = \tilde{\mathrm{h}} \approx \mathrm{h_i}$). Then, inputting $\tilde{\mathrm{h}}$ to $\mathcal{T}_i$, should result in the same predictions for all $m$ predictors associated to $\mathcal{T}_i$.
Since two shallow layers output different representations (i.e., $\mathcal{H}_i(x) \neq \mathcal{H}_j$), the encoder-decoder pair cannot replace the shallow layers for any $\mathcal{H}_j$ where $i \neq j$. Accordingly, after training an initial encoder-decoder pair, FOOL instantiates $n-1$ additional decoders, resulting in a set of separate ${dec_1, dec_2 \dots dec_n}$ decoders, i.e., one for each target backbone. 
\subsection{Image Reconstruction} \label{subsec:imagerecon}
FOOL trains the compression and image reconstruction models in two stages. 
After training the compression model and freezing encoder weights, it separately trains a reconstruction model that maps $\boldsymbol{\hat{\mathrm{y}}}$ to an approximation $\boldsymbol{\tilde{\mathrm{x}}}$ of the original sample $\boldsymbol{{\mathrm{x}}}$.
\subsubsection{Separate Training over Joint Optimization}
We could reduce the distortion $d(\boldsymbol{\mathrm{x}}, \boldsymbol{\hat{\mathrm{x}}})$ with a joint objective for training the reconstruction and compression model. While the resulting models would score higher on the sum of error benchmarks, the added distortion term will result in higher bitrates. Instead, after optimizing the encoder with the objective of SVBI, we freeze the weights (i.e., ``locking'' in the rate performance). Then, we leverage the high mutual information between shallow features and the input to recover presentable approximations (\Cref{subsubsec:machineandhumanperception}). 

Image recovery is closely related to image restoration, such as super-resolution or denoising. The component is exchangeable with the state-of-the-art, 
as it is orthogonal to the compression task. For this work, we select SwinIR~\cite{swinir} due to its relative recency, computational efficiency, and simplicity. 
\subsubsection{Reconstruction Does not Replace Decoders}
Approximations from decoders (\Cref{subsec:encmultisupport}) resulting in (near) lossless prediction would evidence $\hat{\boldsymbol{\mathrm{y}}}$ has sufficient information to reconstruct a sample for the input layers that result in comparable task performance. Hence, after training the encoder, we could replace all decoders with the reconstruction model to approximate the input sample.
Nevertheless, sufficiency may not directly result in lossless prediction since artifacts perturb the reconstructed samples.

We could account for the perturbation by finetuning for a relatively small number of iterations~\cite{frankensplit}. The downside is that operators must maintain, store, and serve different versions of the otherwise identical backbones for each client separately. Worse, they may need to re-train the predictors of the various downstream tasks for each backbone. Instead, we train small decoders that directly map the low-dimensional encoder output to an adequate representation, i.e., FOOL does \textit{not} pass the reconstruction to prediction models for downstream tasks.
There are two advantages to introducing multiple decoders over multiple backbone weights. First, the number of additional weights operators must maintain only scales with supported backbones and not the number of backbone-task pairs. 
Second, the small decoder weights incur considerably less training and storage overhead than the weights of massive backbones.
\subsection{Loss Functions} \label{subsubsec:trainingobj}
FOOL’s training algorithm starts with extracting the shallow layers of a particular detection model (teacher). Then, it freezes the encoder and trains newly initialized decoders ${dec_1, dec_2, \dots dec_n}$ using the corresponding teacher models ${\mathcal{H}_1, \mathcal{H}_2, \dots \mathcal{H}_m}$ (i.e., target shallow layers).  
\subsubsection{Rate-Distortion Loss Function for SVBI} \label{subsubsec:svbirdloss}
To simplify the loss expression, we treat the components related to keypoints as part of $h_a$ if we exclusively use it as a hint for the side-information network. Alternatively, it may be used as the final block of $g_a$ before quantizing and entropy coding the latent.
Analogous to the SVBI training objective \cite{frankensplit, es1},  we have a parametric analysis transform $g_a(x; \theta)$ that maps $x$ to a latent vector $z$. Then, a quantizer $Q$ discretizes $z$ to $\hat{z}$ for lossless entropy coding. Since we rely on HD (\Cref{fig:hd}) as a distortion function, the parametric synthesis transforms $g_{s}(\hat{x};\phi)$ that maps $\hat{y}$ to an approximation of a representation $\tilde{h}$. As introduced in~\cite{fp}, we apply uniform quantization $Q$, but replace $Q$ with continuous relaxation by adding uniform noise  $\eta \thicksim \mathcal{U}(-\frac{1}{2}, \frac{1}{2})$ during training for gradient computation.

Without a hyperprior, the loss is:
\input{embed/equations/var_img_comp_obj_nohyper}
With side information, we condition on a hyperprior, such that each element $\hat{\mathrm{y}}_i$ is now modeled as a Gaussian with its own mean and standard deviation:
\input{embed/equations/modelling_em_latent}
where $\boldsymbol{{\mathrm{z}}} = h_a(\boldsymbol{\hat{\mathrm{y}}};\theta_h)$ and $\boldsymbol{\hat{\mu}}, \boldsymbol{\hat{\sigma}} = h_s(\boldsymbol{\tilde{\mathrm{z}}};\phi_h)$.
The final loss function results in the following:
\input{embed/equations/var_img_comp_obj_whyper}
For the distortion term, we use the sum of squared errors between the shallow layer (teacher) representation and the compressor (student) approximation, i.e., $\operatorname{sse}(\boldsymbol{\mathrm{h}},\tilde{\boldsymbol{\mathrm{h}}})$.
\subsubsection{Mapping Encoder Output to Target Representations} \label{subsubsec:reconloss}
After training the first $enc, dec_1$ pair, FOOL freezes $enc$ weights, i.e., only applying the distortion term of the loss in \Cref{eq:compobjwithsideinfo}, for subsequent decoders  $dec_{2}, dec_{3} \dots dec_{n}$ (\Cref{subsec:encmultisupport}).
%
Lastly, FOOL treats the reconstruction model $rec$ as a decoder and assigns the identity function as its teacher, i.e., $\mathcal{H}_{rec}(x) = x$. Unlike for other decoders, the target representation must be human-interpretable. Hence, we train the decoder for image reconstruction using the \textit{Charbonnier Loss}~\cite{charbonnier}
\input{embed/equations/charbonnier} 
where $\epsilon$ is a small constant we set as $2 \cdot 10^{-3}$.
It is out of this work's scope to exhaustively evaluate image restoration methods. Rather, the focus is to provide empirical evidence for the claims in \Cref{subsubsec:genshallowfeatures}. We simply found that despite performing comparable to other sums of error losses on benchmark metrics, using Charbonnier results in more stable training.

%% file: embed/media/fig_arch_components_org.tex
\begin{figure}[htb]
    \centering
    \includegraphics[width=\columnwidth]{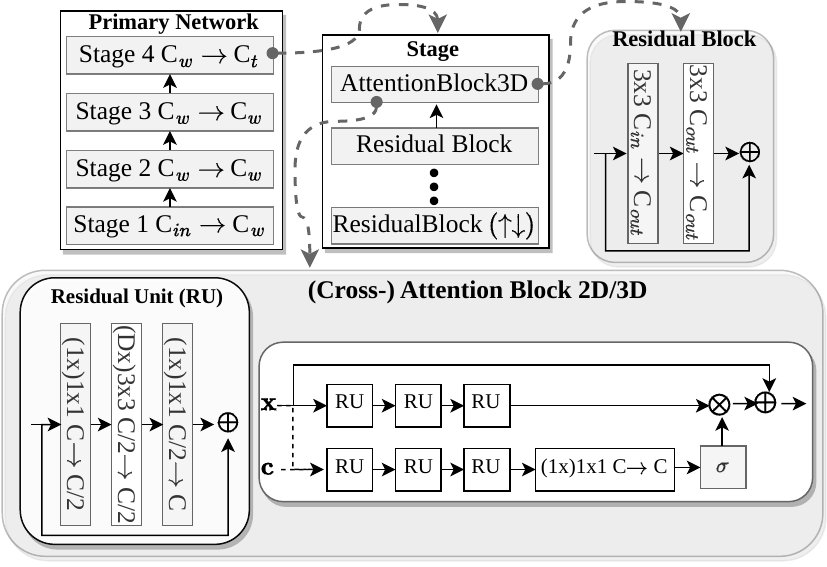}
    \caption{Network Organization and Components}
    \label{fig:archorgcomp}
\end{figure}

%% file: embed/media/fig_leftoverdep_vs_keypoints-whmap.tex
\begin{figure}[htb]
    \centering
    \includegraphics[width=\columnwidth]{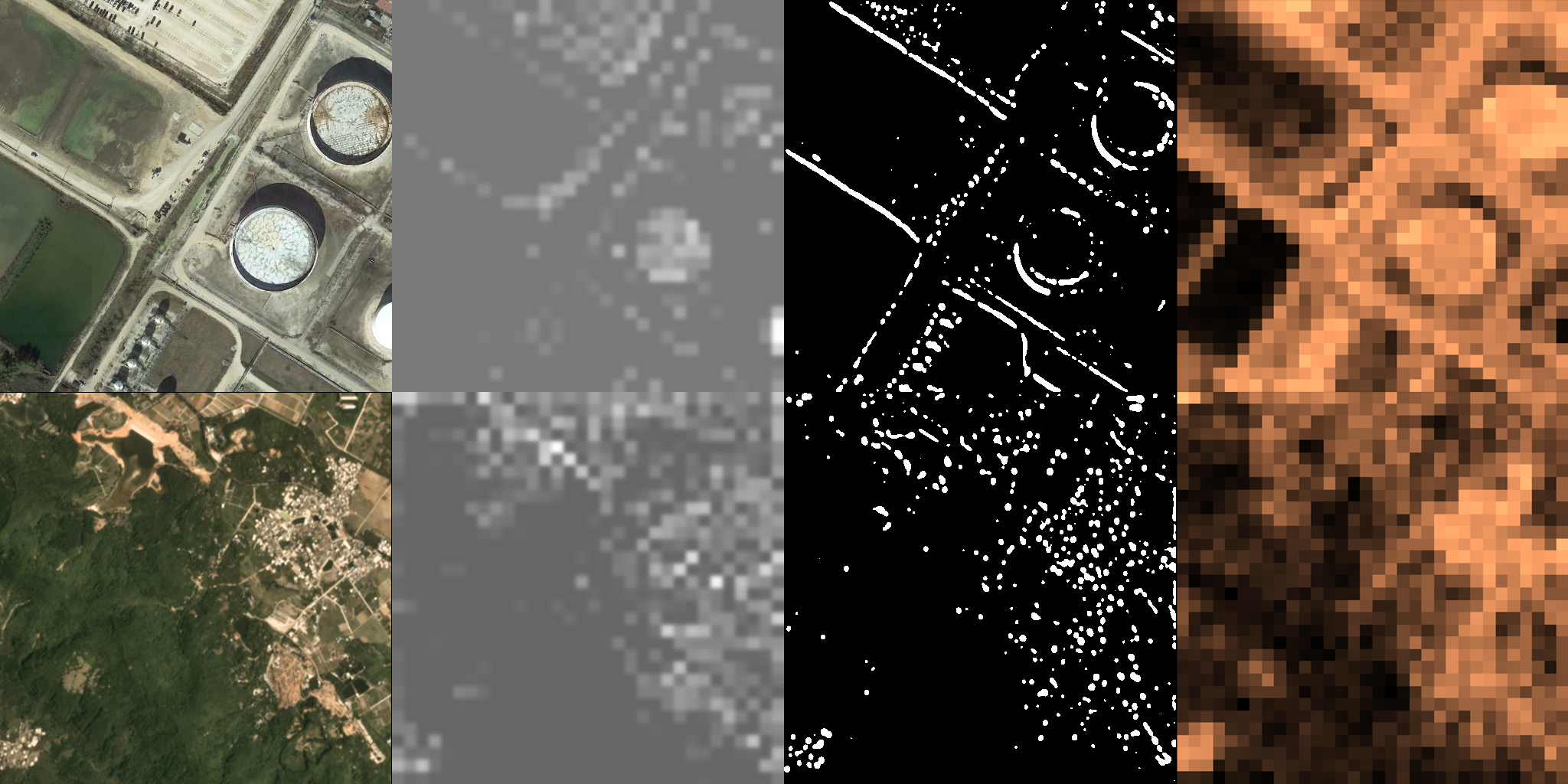}
    \caption{Leftover Spatial Dependencies (middle left), Keypoints (middle right), Entropy Heatmap (right)}
    \label{fig:leftovervskeypoints}
\end{figure}

%% file: embed/equations/sideinfo_with_kps.tex
\begin{align}
    \boldsymbol{\hat{y}} &= Q(g_a({\boldsymbol{x};\theta})) \\
    \boldsymbol{y}_{kp} &= f_{ds}(k(\boldsymbol{x}) \odot \boldsymbol{x};\omega_{kp}) \\
    \boldsymbol{y}_{ca} &= a_c(\boldsymbol{y}_{kp}, \omega_{ca}) \\
    \hat{\boldsymbol{z}} &= Q(h_a(f_{ca}(\boldsymbol{y}_{kp}, \omega_{ca}); \theta_h)) \\
    p_{\hat{\boldsymbol{y}} \mid \hat{\boldsymbol{z}}}(\hat{\boldsymbol{y}} \mid \hat{\boldsymbol{z}}) &\leftarrow h_s\left(\hat{\boldsymbol{z}} ; \phi_h\right) \\
    \tilde{\boldsymbol{h}} &= g_s(\hat{\boldsymbol{y}};\phi)
\end{align}

%% file: embed/media/fig_fool_comp_arch.tex
\begin{figure}[htb]
    \centering
    \includegraphics[width=\columnwidth]{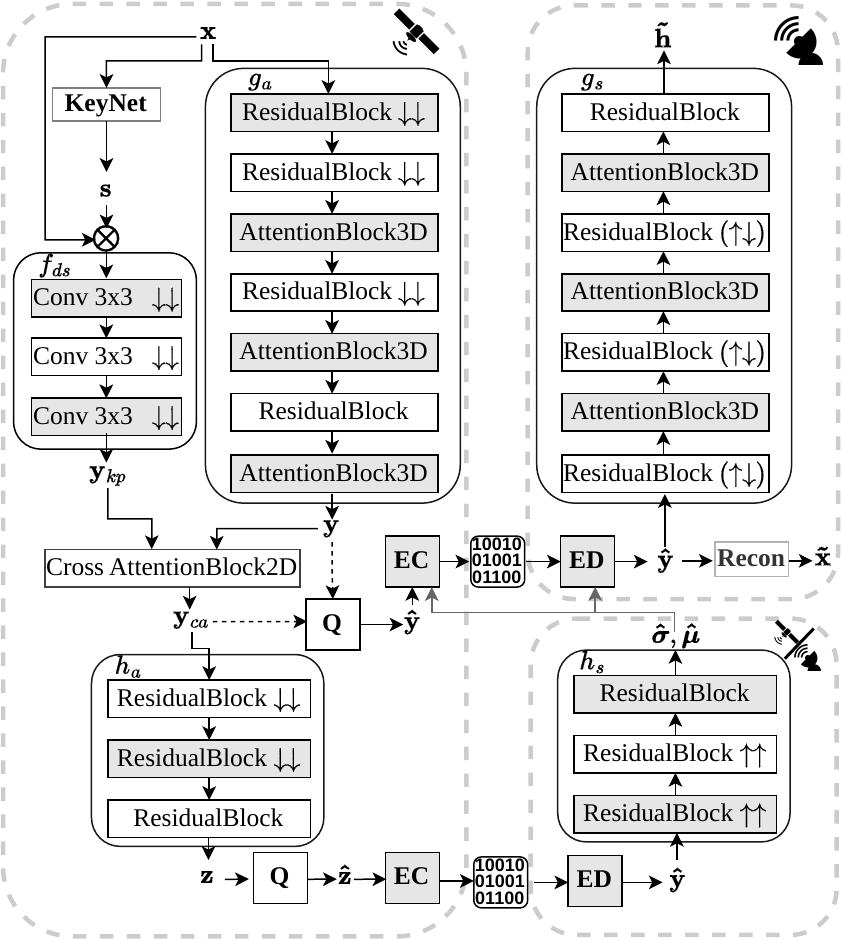}
    \caption{The FOOL's Compression Architecture}
    \label{fig:foolcomparch}
\end{figure}

%% file: embed/equations/bbpairs.tex
\begin{equation}
\mathcal{M}_{f} ={ (\mathcal{H}_1, \mathcal{T}_1), (\mathcal{H}_2, \mathcal{T}_2) \dots (\mathcal{H}_n, \mathcal{T}_n) } \label{eq:bbpairs}
\end{equation}

%% file: embed/equations/var_img_comp_obj_nohyper.tex
\begin{equation}
\mathbb{E}_{\boldsymbol{x} \sim p_{\boldsymbol{x}}} D_{\mathrm{KL}}\left[q \| p_{\tilde{\boldsymbol{\mathrm{y}}} \mid \boldsymbol{x}}\right]=\mathbb{E}_{\boldsymbol{x} \sim p_{\boldsymbol{x}}} \mathbb{E}_{\tilde{\boldsymbol{\mathrm{y}}} \sim q}-\mathrm{log}\;p(x|\boldsymbol{\tilde{\mathrm{y}}}) -\mathrm{log}\; p(\boldsymbol{\tilde{\mathrm{y}}})]\label{eq:compobjnosideinfo} 
\end{equation}

%% file: embed/equations/modelling_em_latent.tex
\begin{equation}
p_{\tilde{\boldsymbol{\mathrm{y}}} \mid \tilde{\boldsymbol{\mathrm{z}}}}\left(\tilde{\boldsymbol{\mathrm{y}}} \mid \tilde{\boldsymbol{\mathrm{z}}}, \phi_h\right)=\prod_i\left(\mathcal{N}\left(\mu, \tilde{\sigma}_i^2\right) * \mathcal{U}\left(-\frac{1}{2}, \frac{1}{2}\right)\right)\left(\tilde{y}_i\right) \label{eq:memlatent}
\end{equation}

%% file: embed/equations/var_img_comp_obj_whyper.tex
\begin{align}
\mathbb{E}_{\boldsymbol{\mathrm{x}} \sim p_{\boldsymbol{\mathrm{x}}}} D_{\mathrm{KL}}\left[q \| p_{\boldsymbol{\mathrm{\tilde{y}}}, \boldsymbol{\mathrm{\tilde{z}}} \mid \boldsymbol{\mathrm{x}}}\right] &= \mathbb{E}_{\boldsymbol{\mathrm{x}} \sim p_{\boldsymbol{\mathrm{x}}}} \mathbb{E}_{\boldsymbol{\mathrm{\tilde{y}}}, \boldsymbol{\mathrm{\tilde{z}}} \sim q}\left[\log q(\boldsymbol{\mathrm{\tilde{y}}}, \boldsymbol{\mathrm{\tilde{z}}} \mid \boldsymbol{\mathrm{x}}) \right. \nonumber \\
& \left. -\log p_{\boldsymbol{\mathrm{x}} \mid \boldsymbol{\mathrm{\tilde{y}}}}(\boldsymbol{\mathrm{x}} \mid \boldsymbol{\mathrm{\tilde{y}}}) -\log p_{\boldsymbol{\mathrm{\tilde{y}}} \mid \boldsymbol{\mathrm{\tilde{z}}}}(\boldsymbol{\mathrm{\tilde{y}}} \mid \boldsymbol{\mathrm{\tilde{z}}})\right. \nonumber \\
& \left. -\log p_{\boldsymbol{\mathrm{\tilde{z}}}}(\boldsymbol{\mathrm{\tilde{z}}})\right] \label{eq:compobjwithsideinfo}
\end{align}

%% file: embed/equations/charbonnier.tex
\begin{equation}
\mathcal{L}_{rec} = \sqrt{\|\boldsymbol{\mathrm{x}} - rec(enc(\boldsymbol{\mathrm{x}}))\|^2 + \epsilon^2}
\end{equation}

%% file: sections/evaluation/evaluation.tex
\section{Evaluation} \label{sec:eval}
\input{sections/evaluation/preamble}
\input{sections/evaluation/compression}
\input{sections/evaluation/system}

%% file: sections/evaluation/preamble.tex
\subsection{Experiment Design and Methodology} \label{sec:evalpreamble}
Our experiments reflect our aim to determine (i) the compression performance on aerial and satellite imagery without relying on prior knowledge and (ii) the feasibility of orbital inference.
\subsubsection{Testbed} \label{subsubsec:testbed} 
We benchmark~\cite{e2ebench} on an analytic and trace-driven~\cite{faassim} simulation based on results from a physical testbed with hardware summarized in \Cref{tab:hwconf}. 
The power consumption is capped at $15$W for the entire testbed.
\input{embed/tables/hardware}
Our simulation replicates a configurable CubeSat by imposing energy, memory, and bandwidth constraints. To simulate the downlink bottleneck with varying link conditions, parameterize link conditions and data volume (\Cref{subsec:oec}) using real-world missions~\cite{dovehsd, worldview3, sentinel3, landsat8specs} as summarized in \Cref{tab:missions}. 
\input{embed/tables/missions}
 Due to the orthogonality of compression to systems-related challenges in OEC, we argue a focused simulation yields more insight results than running a full OEC simulator (e.g., \cite{denby2020}). 
%
Our intention is for FOOL to facilitate OEC as an auxiliary method. Therefore, we demonstrate the bitrate reduction and resource usage trade-off for various configurations representing the heterogeneity of available compute resources and nanosatellite constellations. 
\subsubsection{Third-Party Detection Models \& Target Tasks} \label{subsubsec:detectors}
FOOL derives the basic approach to accommodate multiple backbones with a single encoder (\Cref{subsec:encmultisupport}) from FrankenSplit~\cite{frankensplit}. 
Foundational models (i.e., feature extractors or backbones) are interchangeable third-party components in SVBI. To complement previous work (\Cref{subsec:colinference}) and further show the flexibility of SVBI, we focus on modern YOLO variants~\cite{ultralytics}. 
\Cref{fig:backbonesdetect} illustrates the pipeline to represent third-party detectors we prepare before evaluating codecs. 

\input{embed/media/fig_backbones_detection}
While the work in \cite{frankensplit} did not explicitly evaluate object detection tasks, the support for two-stage detectors follows from the codec sufficiently approximating the representation of the feature extractor (i.e., the first stage). However, it is not apparent whether the general SVBI framework yields gains over image codecs when the targets are one-stage detectors. Therefore, to replicate a representative service for inference on aerial or satellite imagery, we apply simple transfer learning on open-source weights~\cite{ultralytics} for YOLOv5 and YOLOv8.
Image codecs pass a sample $dec(enc(\boldsymbol{\mathrm{x}})) = \boldsymbol{\hat{\mathrm{x}}}$ to the input layers of a target model. Feature codecs (i.e., SVBI methods) skip the shallow layers and pass $dec(enc(\boldsymbol{\mathrm{x}})) = \boldsymbol{\hat{\mathrm{h}}}$ to the deeper layers. Detection models with the same architecture share the frozen shallow layers (i.e., layers until the first non-residual connection). 
We associate one task as outlined by the test labels for each of the three dataset separately. The tasks represent varying mission conditions.
DOTA-2~\cite{dota} for a more coarse-grained aerial task with comparatively lower Ground Sample Distance (GSD) and larger objects. SpaceNet-3~\cite{spacenet} for urban tasks (e.g., for traffic control) with high image resolutions. Lastly, xView~\cite{xview} for disaster response systems where detection models rely on fine-grained details.  Lastly, image reconstruction is treated distinctly as single task for the reconstruction model, and not the detection pipeline, by combining the images from the three test sets.

To simplify the already intricate evaluation setup and to ease reproducibility, we deliberately refrain from more refined transfer learning methods. We merely require detection models with mAP scores that are moderately high to determine whether a codec can preserve fine-grained details for EO tasks on satellite imagery. For each architecture, we jointly finetune the deeper layers and train separate predictors that achieve around 35-65\% mAP@50. 
\subsubsection{Training \& Implementation Details} \label{subsubsec:impldetails}
To demonstrate that FOOL can handle detection tasks without relying on prior information (\Cref{subsec:limexistingcodecs}), we do not optimize the compression model with the training set of the prediction tasks (i.e., DOTA-2, SpaceNet-3, xView). Instead, 
we curate other aerial and satellite datasets~\cite{airbus, aerialcrowd, spacenet8, spacenet7, spacenet6, rareplanes, hrplanesv2, floodnet} that cover region and sensor diversity. SVBI does not rely on labels, i.e., replacing the curation with any diverse enough dataset from satellite imagery providers (e.g., Google Earth Engine) should be possible. 

We train one separate compression mode for each third-party detector using the shallow layers as teachers and verify whether the rate-distortion performance is comparable. Then, we freeze the encoder of the compression model for YOLOv5-L and discard all other encoders.
Lastly, we freeze the remaining encoder’s weights and (re-)train the separate decoders to demonstrate clients may request inference on variations (YOLOv5-M) or newer models as they emerge (YOLOv8).

We fix the tile resolution to $512 \times 512$ during training. We load samples as a video sequence for FOOL by grouping tiles from the same image in partitioning order with random transformations to fill any remaining spots. After training, the tensor shape (i.e., the number of tiles and the spatial dimensions) may vary for each separate sample. 
%
We use PyTorch~\cite{pytorch}, CompressAI~\cite{begaint2020compressai}, and pre-trained detection models from Ultralytics~\cite{ultralytics}. To ensure reproducibility, we use torchdistill~\cite{matsubara2021torchdistill}. 
%
We use an Adam optimizer~\cite{kingma2014adam} with a batch size of $8$ and start with an initial learning rate of $1 \cdot 10^{-3}$, then gradually lower it to $1 \cdot 10^{-6}$ with an exponential scheduler. We first seek a weight for the rate term in \Cref{eq:compobjwithsideinfo} that results in lossless prediction with the lowest (best) bpp. Then, we progressively increase the term weight to evaluate trade-offs between rate and predictive loss. 
\subsubsection{Datasets Preparation} \label{subsubsec:datapreprocess}
The train sets for third-party detectors and the train sets for the compression models are strictly separated. However, we create square tiles for all datasets by partitioning the images with a configurable spatial dimension and applying 0-padding where necessary. 
We extract bands from samples corresponding to RGB and convert them to 8-bit images, as to the best of our knowledge, there are no widespread open-source foundational models for detection with multispectral data yet. 
To ease direct comparisons, we convert the network detection labels of SpaceNet-3 by transforming the polygonal chains into bounding boxes. Lastly, since there are no publicly available labels for the xView and SpaceNet test sets, we create a 9:1 split on the train set.
\subsubsection{Compression Performance Measures} \label{subsubsec:compmeasure}
To evaluate how codecs impact downstream task performance, we measure \emph{Predictive Loss} as the drop in mean Average Precision (mAP) by inputting decoded samples. We regard a configuration to result in \emph{lossless prediction} if there is less than 1\% difference in expected mAP@50. We confirm the observations from \cite{frankensplit} where the initial teacher only negligibly affects compression performance, and the predictive loss by a codec is comparable across target models (i.e., the retained information in shallow layers is similar across YOLO variations). Hence, for brevity, we aggregate the compression performance for each task separately, taking the highest predictive loss incurred on a detection model. 
We train the image reconstruction model using the same configurations as \cite{swinir}, and compare it with LIC models using common measures (PSNR, MS-SSIM, LPIPS~\cite{lpips}).
\subsubsection{Baselines} \label{subsubsec:baselines}
We consider seminal work for image codecs as baselines with available open-source weights. Factorized Prior (FP)~\cite{fp} as a relatively small model without side information. (Mean-)scale hyperprior (SHP, MSHP)~\cite{shp, jahp} for drawing comparisons to side information in LIC, and Joint autoregressive and hierarchical priors (JAHP)~\cite{jahp} that further improves compression performance with an autoregressive context model. Lastly, TinyLIC~\cite{tinylic} represents recent work on efficient LIC design with state-of-the-art rate-distortion performance. 
\input{embed/tables/n_codecs_params_macs}
\Cref{tab:paramsmacsummary} summarizes parameter distributions between encoder and decoder components from LIC models. 

To draw comparisons to existing work on SVBI, we combine FrankenSplit~\cite{frankensplit} and the Entropic Student~\cite{es2} as a single baseline (BSVBI). We utilize FrankenSplit’s more efficient architecture design since it outperforms the latter without relying on finetuning the deeper layers. The encoder consists of stacked residual blocks (\Cref{subsec:modelbuildingblocks}), and the decoder is instantiated from a YOLOv5+ blueprint (C3 blocks~\cite{ultralytics}). We scale the capacity of BSVBI by including side information from LIC (MSHP) and increasing the width and depth to match the various FOOL configurations.
We train FOOL and BSVBI with the same dataset and training parameters (\Cref{subsubsec:impldetails}).

%% file: embed/tables/hardware.tex
\begin{table}[htb]
\centering
\caption{Testbed Device Specifications}
\label{tab:hwconf}
\begin{tabularx}{\columnwidth}{@{\extracolsep{\fill}} 
l 
>{\raggedleft\arraybackslash} p{0.313\linewidth}
>{\raggedleft\arraybackslash} p{0.313\linewidth}
}
\makecell[c]{Device}  & \makecell[c]{CPU} & \makecell[c]{GPU} \\ \toprule
\rowcolor[HTML]{EFEFEF} 
Ground Server    & 16x Ryzen @ 3.4 GHz & RTX 4090          \\
Edge (Nano Orin) & 6x Cortex @ 1.5 GHz & Amp. 512 CC 16 TC \\
\rowcolor[HTML]{EDEDED} 
Edge (TX2)       & 4x Cortex @ 2 GHz   & Pas. 256 CC       \\
Edge (Xavier NX) & 4x Cortex @ 2 GHz   & Vol. 384 CC 48 TC \\ \bottomrule
\end{tabularx}
\end{table}

%% file: embed/tables/missions.tex
\begin{table}[htb]
\centering
\caption{Constellation Link Conditions}
\label{tab:missions}
\begin{tabularx}{\columnwidth}{@{\extracolsep{\fill}} 
@{}l@{ }@{ }
l @{ }@{ }@{ }
l@{ }@{ }
>{\raggedleft\arraybackslash} p{0.11\linewidth}@{  }@{  }
>{\raggedleft\arraybackslash} p{0.07\linewidth}@{  }@{  }
>{\raggedleft\arraybackslash} p{0.138\linewidth} }
\makecell[l]{Oper.} & \makecell[l]{Constellation} & \makecell[l]{Link} & 
\makecell[r]{Rate\\(Mbps)} & \makecell[r]{Pass\\(s)} & \makecell[r]{Data Per\\Pass (GB)}
\\ \toprule
\rowcolor[HTML]{EFEFEF} 
Planet & Dove (3P/B13) & HSD 1         & 160  & 510 & 12.0  \\
Maxar  & WorldView     & WorldView-3   & 1200 & 600 & 90 \\
\rowcolor[HTML]{EFEFEF}
ESA    & Copernicus    & Sentinel 3A/B & 560  & 600 & 40.0  \\
\rowcolor[HTML]{FFFFFF}
NASA & Landsat & Landsat 8 & 440 & 120 & 39.6 \\
\hline
\end{tabularx}
\end{table}

%% file: embed/media/fig_backbones_detection.tex
\begin{figure}[htb]
    \centering
    \includegraphics[width=\columnwidth]{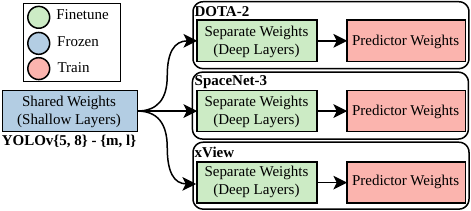}
    \caption{Detection Pipeline for Evaluation}
    \label{fig:backbonesdetect}
\end{figure}

%% file: embed/tables/n_codecs_params_macs.tex
\def\myParColWidth{0.165}
\begin{table}[ht]
\centering
\caption{Summary of Codec Parameter Distribution}
\label{tab:paramsmacsummary}
\begin{tabularx}{\columnwidth}{@{\extracolsep{\fill}} 
l 
>{\raggedleft\arraybackslash}p{\myParColWidth\linewidth}
>{\raggedleft\arraybackslash}p{\myParColWidth\linewidth}
>{\raggedleft\arraybackslash}p{\myParColWidth\linewidth}
>{\raggedleft\arraybackslash}p{\myParColWidth\linewidth}}
Codec   & Pars. Total & Pars. Enc. & Pars. Dec. & Shared \\ 
\toprule 
FOOL-L  & 4.97M        & 1.19M       & 4.06M       & 0.28M         \\
\rowcolor[HTML]{EFEFEF} 
FOOL-M  & 4.33M        & 0.69M       & 3.83M       & 0.16M         \\
FOOL-S  & 3.91M        & 0.35M       & 3.66M       & 0.08M         \\
\rowcolor[HTML]{EFEFEF} 
SVBI-L  & 4.99M        & 1.25M       & 4.39M       & 0.65M         \\
SVBI-M  & 4.35M        & 0.72M       & 3.91M       & 0.29M         \\
\rowcolor[HTML]{EFEFEF} 
SVBI-S  & 3.95M        & 0.38M       & 3.71M       & 0.16M         \\
FP      & 7.03M        & 3.51M       & 3.51M       & 0.02M         \\
\rowcolor[HTML]{EFEFEF} 
MSHP    & 17.56M       & 14.06M      & 11.66M      & 8.15M         \\
JAHP    & 25.50M       & 21.99M      & 19.60M      & 16.10M        \\
\rowcolor[HTML]{EFEFEF} 
TinyLIC & 28.34M       & 21.23M      & 19.16M      & 12.05M        \\ 
\bottomrule
\end{tabularx}
\end{table}

%% file: sections/evaluation/compression.tex
\subsection{Rate Trade-off with Predictive Loss} \label{subsec:compperfmachine}
We report the predictive loss as a percentage point difference using mAP@50 on foundational detection models.
\subsubsection{Comparison to Image Codecs} \label{subsubsec:complic}
\Cref{plot:compimagerdtask} illustrates the trade-off between bpp (left is better) and predictive loss (top is better) for LIC models on each task separately.
\input{embed/media/plot_comp_image_perf_machine}
We primarily focus on how FOOL compares to existing SVBI to draw new insights from the novel additions and confirm that our results on aerial and satellite imagery datasets with one-stage detectors are consistent with previous findings on standardized terrestrial datasets~ \cite{frankensplit, es1, es2}. 
 %
\subsubsection{Comparison to Feature Codecs} \label{subsubsec:compfeat}
\Cref{plot:compfeatrdtask} contrasts the trade-off between bpp and predictive loss for FOOL and BSVBI with progressively increasing sizes (i.e., capacity). 
The efficacy of compressing shallow features is best shown by comparing BSVBI and FOOL to MSHP, as they rely on the same entropy model. The highest quality MSHP model results in about 3-4\% predictive loss for DOTA-2. In contrast, the highest quality BSVBI-S model has 37x fewer encoder parameters but results in half the bitrate with no predictive loss.
\input{embed/media/plot_comp_feature_perf_machine}
Despite BSVBI demonstrating strong compression performance, FOOL significantly outperforms BSVBI across all configurations. FOOL-S has a ~51\% lower bitrate for configurations with lossless prediction than the comparatively large BSVBI-L. Relative to the FOOL model with matching capacity (FOOL-L), BSVBI-L has twice the bitrate.  
\subsubsection{Ablation Study} \label{subsubsec:ablation}
We may consider BSVBI an ablation, as FOOL extends BVSBI’s architecture by placing 3D attention layers between the residual blocks and a cross-attention layer to include context. The auxiliary networks $h_a$ and $h_s$ (\Cref{subsec:contextforfcomp}) are identical for FOOL and SVBI, i.e., three stacked residual blocks. Additionally, we perform ablation studies to assess by-component improvement and summarize the results for lossless predictions in \Cref{tab:ablation}.

\input{embed/tables/ablation}
The NITA models include the keypoint context without the inter-tile attention (ITA) layers. Analogous to BSVBI, we replace attention layers with residual blocks and match corresponding model sizes by increasing the depth and width of NITA models. 
NKPC-Ablation drops components for embedding keypoints, i.e., it only includes the IT attention layers.

The results show that relative to BSVBI, the task-agnostic context component contributes considerably more to rate reductions 
than the ITA layers
that leverage inter-tile spatial dependencies. 
Still, we argue that the NTI-layers fulfill their purpose, to synergize with the partitioning strategy that maximizes processing throughput (\Cref{subsec:sysperf}).
\subsection{Image Reconstruction Quality} \label{subsec:compperfhuman}
We aim to demonstrate the feasibility of recovering presentable images from the compressed latent space of shallow features. We average results on DOTA-2, SpaceNet-3, and xView to reduce the bloat of reporting similar values summarize the results in \Cref{tab:imagereconeval}. For transparency, we exclusively select samples from the lower quartile across all measures to qualitatively showcase the reconstruction.
 
\input{embed/tables/recon_perf}
HQ refers to the weights with the highest available quality, and MQ refers to mid-quality weights that roughly match FOOL in PSNR. FOOL-FT finetunes the reconstruction model for an additional $2.5 \cdot 10^{5}$ iterations using LPIPS~\cite{lpips}. 
Unsurprisingly, the LIC models achieve significantly better scores across all reconstruction measures (i.e., PSNR, MS-SSIM, and LPIPS). The advantage of FOOL is that it has a considerably lower bitrate with no predictive loss on tasks for which it had no prior information.  Nonetheless, the results are considerably more interesting when contrasting FOOL to LIC models with mid-quality weights. Notice how FOOL matches reconstruction measures at no predictive loss and a 46-77\% \emph{lower} bitrate. Note that we did not find that the dataset significantly influences rate-distortion performance, except for a slight reduction in predictive loss (verified by training an FP model on the curation using the same setup as in \cite{begaint2020compressai}). Compression is a low-level vision task that generalizes well but may lack domain specificity when applying a standard rate-distortion reconstruction loss. In other words, the objective is the decisive difference between SVBI and LIC models. 
\input{embed/media/fig_qual_analysis}
To provide some intuition to the LPIPS measure, we select an image where FOOL-FT achieves considerably lower PSNR than TinyLIC and contrast the results in \Cref{fig:qual_analysis}.
Notice that the quality differences are most visible with fine-grained details, i.e., compared to TinyLic-HQ, TinyLic-MQ has a noticeable blur with some shadows completely missing in the bottom left. FOOL-FT preserves such details, despite lower PSNR, and this increase in perceptual quality is reflected in the LPIPS score.
\input{embed/media/fig_qual_analysis_pt2}
\Cref{fig:qual_analysis_pt2} further visualizes the potential of reliably recovering fine-grained from the compressed latent space.
Naturally, it should be possible to finetune the TinyLic-MQ to improve perceptual quality analogous to FOOl-FT. However, TinyLIC is still a significantly costlier model, with a worse rate and prediction performance. 
More pressingly, we stress that the \emph{reliability} of a restoration model is bound by the available signals in the compressed latent space. Accordingly, we deliberately avoid generative models that prioritize realism over structural integrity. 
Prioritizing realism over reliability defeats the primary purpose of image restoration, i.e., intervention by human experts in critical EO applications. A model outperforming experts does not imply that predictions may inexplicably be false. In particular, where human cost is involved (e.g., disaster warning or relief~\cite{disasterwarning}) it is paramount that experts can trust the codec to not include extrapolated elements to an image.  

We argue that our results adequately underpin the statements in \Cref{subsubsec:genshallowfeatures} and \Cref{subsubsec:machineandhumanperception}. In summary, if the salient regions align, compressing for model prediction requires more information than for human observation. Task specificity determines rate savings and not an entity's input interface. Targeting shallow features is minimally task-specific by relaxing the objective for lossless prediction on all possible tasks to those valuable for clients.

%% file: embed/media/plot_comp_image_perf_machine.tex
\begin{figure}[htb]
    \centering
    \includegraphics[width=\columnwidth]{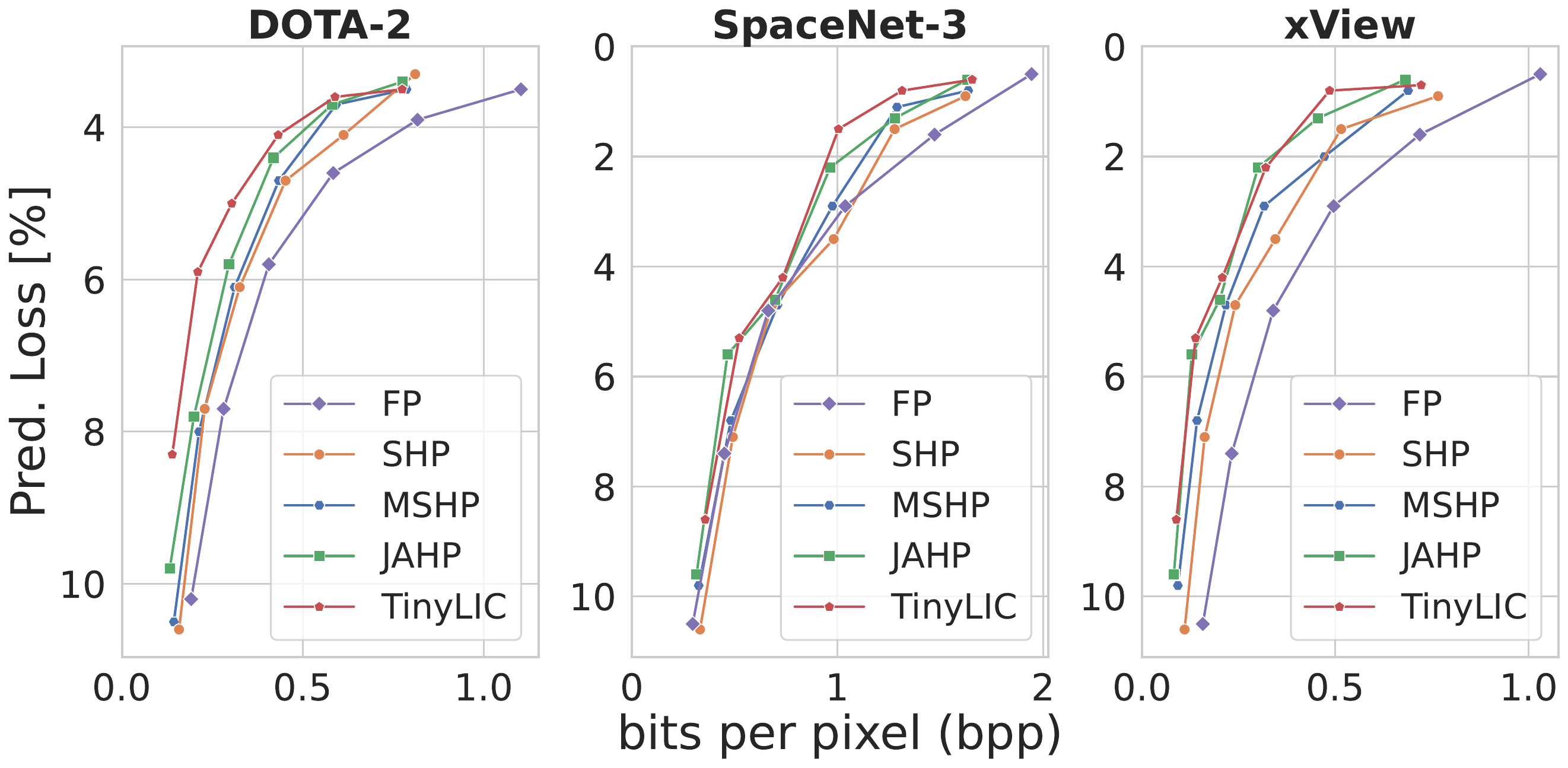}
    \caption{Compression Performance Image Codecs}
    \label{plot:compimagerdtask}
\end{figure}

%% file: embed/media/plot_comp_feature_perf_machine.tex
\begin{figure}[htb]
    \centering
    \includegraphics[width=\columnwidth]{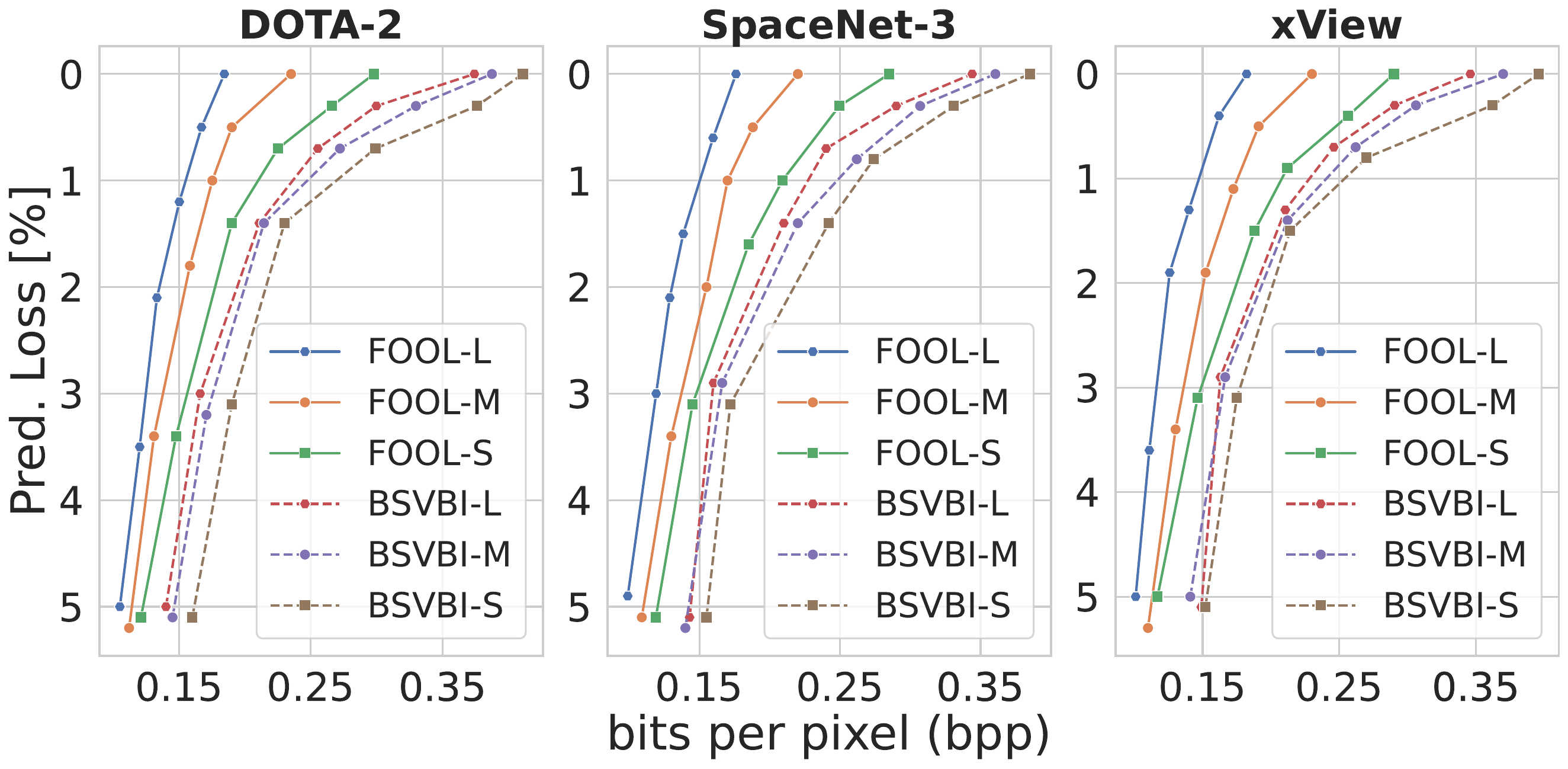}
    \caption{Compression Performance Feature Codecs}
    \label{plot:compfeatrdtask}
\end{figure}

%% file: embed/tables/ablation.tex
\begin{table}[htb]
\centering
\caption{Ablations Comparisons for Lossless Prediction}
\label{tab:ablation}
\begin{tabularx}{\columnwidth}{@{\extracolsep{\fill}} 
l@{}
>{\raggedleft\arraybackslash} p{0.245\linewidth}@{}
>{\raggedleft\arraybackslash} p{0.286\linewidth}@{}
>{\raggedleft\arraybackslash} p{0.22\linewidth} }
Model & DOTA-2 (bpp \textdownarrow \textdownarrow) & SpaceNet-3 (bpp \textdownarrow \textdownarrow) & xView (bpp \textdownarrow \textdownarrow) \\ \toprule
\rowcolor[HTML]{EFEFEF} 
FOOL-L                         & 0.1843 & 0.1760 & 0.1822 \\
FOOL-M                         & 0.2110 & 0.1993 & 0.2032 \\
\rowcolor[HTML]{EFEFEF} 
FOOL-S                         & 0.2389 & 0.223  & 0.2290 \\ \hline
BSVBI-L                        & 0.3622 & 0.3440 & 0.3452 \\
\rowcolor[HTML]{EFEFEF} 
BSVBI-M                        & 0.3775 & 0.3605 & 0.3699 \\
BSVBI-S                        & 0.3889 & 0.3852 & 0.3909 \\ \hline
\rowcolor[HTML]{EFEFEF} 
NITA-L                         & 0.2209 & 0.1993 & 0.2032 \\
NITA-M                         & 0.2287 & 0.2193 & 0.2205 \\
\rowcolor[HTML]{EFEFEF} 
NITA-S                         & 0.2433 & 0.2386 & 0.2407 \\ \hline
NKPC-L                         & 0.2839 & 0.2712 & 0.2768 \\
\rowcolor[HTML]{EFEFEF} 
NKPC-M                         & 0.2926 & 0.2855 & 0.2883 \\
NKPC-S                         & 0.3116 & 0.3029 & 0.3112 \\ \bottomrule
\end{tabularx}
\end{table}

%% file: embed/tables/recon_perf.tex
\begin{table}[htb]
\centering
\caption{Comparison Between Recovery and Image Codecs}
\label{tab:imagereconeval}
\begin{tabularx}{\columnwidth}{@{\extracolsep{\fill}} 
l@{ }@{ }
>{\raggedleft\arraybackslash}p{0.125\linewidth}@{ }@{ }@{ }
>{\raggedleft\arraybackslash}p{0.17\linewidth}@{ }@{ }@{ }
>{\raggedleft\arraybackslash}p{0.12\linewidth}@{ }@{ }@{ }
>{\raggedleft\arraybackslash}p{0.1\linewidth}@{ }@{ }@{ }
>{\raggedleft\arraybackslash}p{0.18\linewidth} }
Model &
PSNR\textuparrow \textuparrow &
MS-SSIM\textuparrow\textuparrow &
LPIPS\textdownarrow\textdownarrow &
BPP\textdownarrow \textdownarrow &
Pred. Loss\textdownarrow\textdownarrow \\ \hline
FOOL       & 36.51                          & 15.43                         & 0.1700                          & 0.1808                         & \textbf{-}               \\
FOOL-FT &
  \cellcolor[HTML]{EFEFEF}35.56 &
  \cellcolor[HTML]{EFEFEF}14.57 &
  \cellcolor[HTML]{EFEFEF}0.1480 &
  \cellcolor[HTML]{EFEFEF}0.1808 &
  \cellcolor[HTML]{EFEFEF}\textbf{-} \\ \hline
FP-HQ      & 43.22                          & 25.07                         & 0.0896                         & 1.0470                          &  1.500                         \\
FP-MQ      & \cellcolor[HTML]{EFEFEF}35.56  & \cellcolor[HTML]{EFEFEF}16.55 & \cellcolor[HTML]{EFEFEF}0.2498 & \cellcolor[HTML]{EFEFEF}0.3200  & \cellcolor[HTML]{EFEFEF}7.508 \\ \hline
MSHP-HQ    & 43.90                          & 25.20                         & 0.0841                         & 1.0370                          & 1.711                          \\
MSHP-MQ    & \cellcolor[HTML]{EFEFEF}36.45  & \cellcolor[HTML]{EFEFEF}16.83 & \cellcolor[HTML]{EFEFEF}0.2361 & \cellcolor[HTML]{EFEFEF}0.2787 & \cellcolor[HTML]{EFEFEF}7.180 \\ \hline
JAHP-HQ    & 43.95                          & 25.17                         & 0.0818                         & 1.0297                         &  1.504                        \\
JAHP-MQ    & \cellcolor[HTML]{EFEFEF}36.61  & \cellcolor[HTML]{EFEFEF}16.95 & \cellcolor[HTML]{EFEFEF}0.2303 & \cellcolor[HTML]{EFEFEF}0.2641 & \cellcolor[HTML]{EFEFEF}6.397 \\ \hline
TinyLIC-HQ & 44.52                          & 25.07                         & 0.0683                         & 1.0473                         &  1.602                        \\
TinyLIC-MQ &
  \cellcolor[HTML]{EFEFEF}37.42 &
  \cellcolor[HTML]{EFEFEF}17.27 &
  \cellcolor[HTML]{EFEFEF}0.2102 &
  \cellcolor[HTML]{EFEFEF}0.2899 &
  \cellcolor[HTML]{EFEFEF}5.499 \\ \hline
\end{tabularx}%
\end{table}

%% file: embed/media/fig_qual_analysis.tex
\begin{figure*}[htb]
    \centering
    \includegraphics[width=\textwidth]{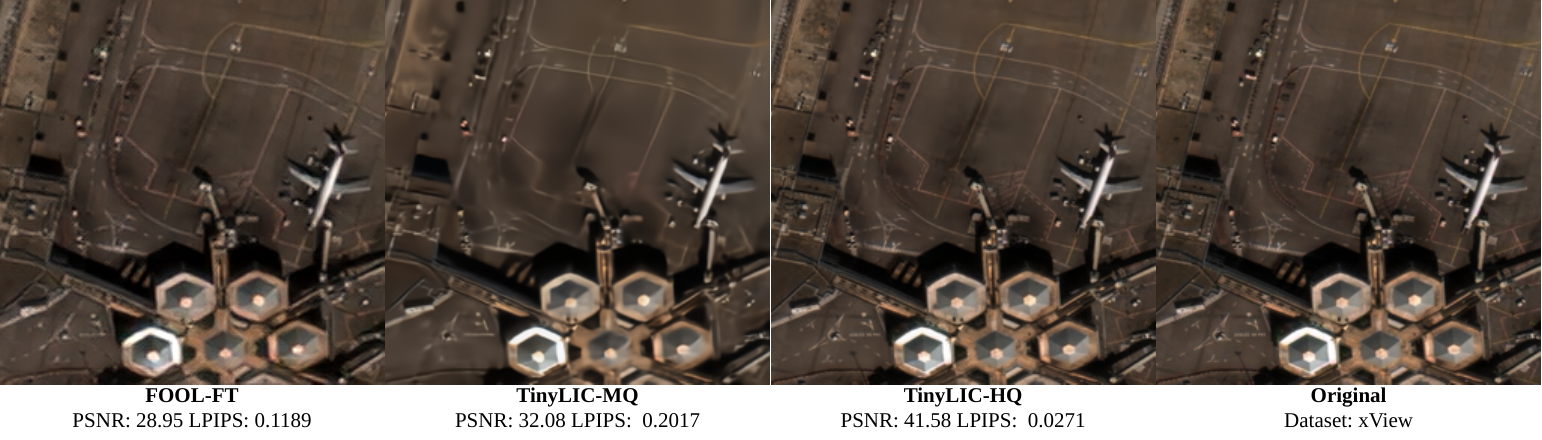}
    \caption{Visual Comparison between FOOL Image Recovery and a State-of-the-Art LIC model}
    \label{fig:qual_analysis}
\end{figure*}

%% file: embed/media/fig_qual_analysis_pt2.tex
\begin{figure*}[htb]
    \centering
    \includegraphics[width=\textwidth]{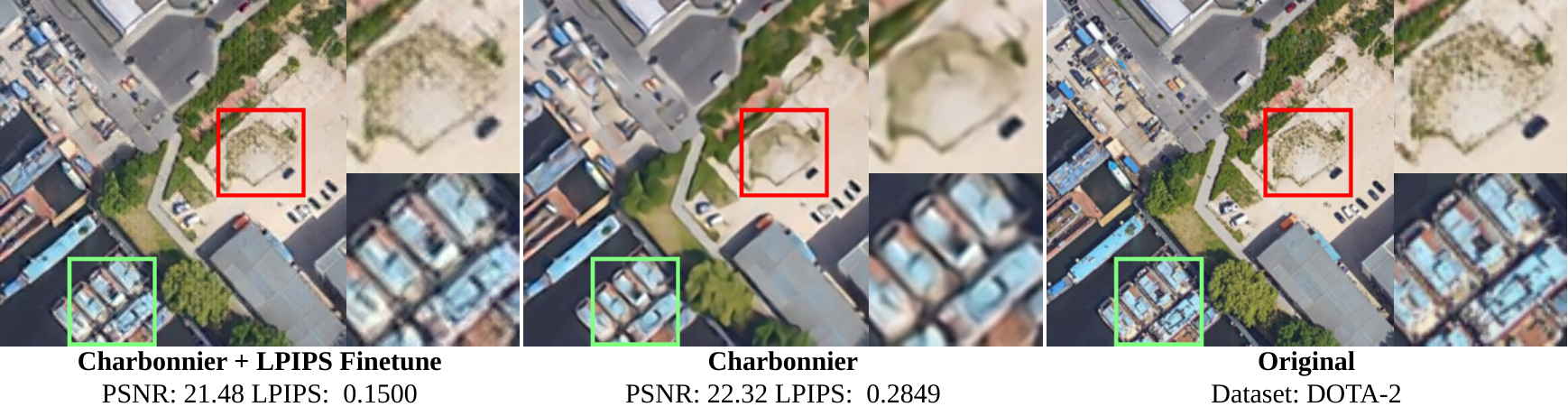}
    \caption{Showcasing Potential of Recovery from Compressed Features with Finetuning for Perceptual Quality using LPIPS}
    \label{fig:qual_analysis_pt2}
\end{figure*}

%% file: sections/evaluation/system.tex
\subsection{System Performance} \label{subsec:sysperf}
The following evaluates FOOL’s resource usage and how well it can address the downlink bottleneck. The methodology resembles how the system aids operators in determining the correct model size for a target device and estimating the increase in data volume relative to bent pipes. We do not apply vendor-specific optimization (e.g., TensorRT) to ensure transparent evaluation and keep the results reasonably platform agnostic. Instead, we instantiate all models dynamically with half-precision in the native PyTorch environment (\texttt{torch  1.14.0} with \texttt{CUDA 11.4.315}).
Image codecs are omitted for conciseness, as even the state-of-the-art for efficient LIC design still runs considerably slower than the largest SVBI models.
\subsubsection{Processing Throughput and Transfer Cost  Reduction} \label{subsubsec:evalinference}
We manually step through parts of the profiler (\Cref{subsec:resourcemaximation}) for evaluation and to show how it estimates gains in downlinkable data volume. 
%
Consider the results from measuring the friction between model sizes and input dimensions on processing throughput in \Cref{plot:throughputmain}. 
\input{embed/media/plot_throughput_main}
Notably, the processing throughput gain of FOOL-S over FOOL-M is significantly higher than FOOL-M over FOOL-L, despite FOOL-M having a comparable size difference to both models.   

\input{embed/tables/optim_config_memory_wthroughput}
\Cref{tab:optimconfigmemorywthroughput} summarizes the configuration that maximizes profiler selection by TCR/s for all models on each device separately. Since bitrate variance is low between DOTA-2, SpaceNet-3, and xView, we average the bpp (\Cref{subsec:compperfmachine}) on the validation sets. The bold Model value indicates the adequate size of each model family on a device, i.e., the model we will deploy to measure data volume downlinking in the following experiments. The bold TCR/s marks the highest overall value for a device, i.e., we can expect applying FOOL over BSVBI to result in considerably more downlinkable data on all devices. However, due to keypoint extraction and the ITA layers, FOOL's processing throughput is slower than that of BVSBI. The overhead is particularly punishing for the most constrained device (i.e., the previous-generation TX2), where BSVBI-M has slightly higher TCR/s than FOOL-M despite the latter's significantly better compression performance. Moreover, the profiler selects FOOL-S over FOOL-M/-L for the low-end current generation Orin Nano and high-end last-gen past generation NX. Conversely, the profiler decides on the mid-sized model for BSVBI across all devices despite FOOL's compression performance scaling better.

Still, we argue that the results accentuate the findings from \Cref{subsubsec:compfeat}.
Notice the contrast between TX2 and Nano Orin. One hardware generation was sufficient for the lowest-end device in the Jetson lineup to see a \emph{threefold} increase in TCR/s on FOOL-L over the last-generation midrange device. Thus, it is reasonable to claim that FOOL can (i) adequately leverage the current rapid progression of energy-efficient hardware improvement (i.e., with FOOL-M, L, and potentially larger variants)  and (ii) is flexible enough to be deployed on more constrained devices using the small FOOL-S that still achieve substantial rate reduction. 
\subsubsection{Model Inference with Concurrent Task Execution} \label{subsubsec:evalinferencewithconcurrent}
The following examines the claim in \Cref{sec:conctask}, i.e., whether FOOL’s compression pipeline can offset the runtime overhead of entropy coding. In other words, we evaluate whether interference between concurrent GPU and CPU-bound processes is negligible enough. We assume the worst case for interference, i.e., the CPU-bound processes constantly run concurrently by keeping them busy from an additional data stream when necessary. As all three devices have multicore CPUs and a dedicated GPU, we report results on the Nano Orin due to space constraints. 

\input{embed/tables/entropycoderperfwbaselines}
\Cref{tab:entropycodingfiles} summarizes the results from running the entire compression pipeline with concurrent task execution using the configurations that maximize TCR/s from \Cref{tab:optimconfigmemorywthroughput}. The bold values in the TCR/s dec. column indicates the size with the highest decrease.
A file includes all model artifacts output by the neural codec's DNN components for a single tile, i.e., the pipeline still needs to entropy code them to match the bpp in TCR/s calculations. 
%
File size refers to the storage requirements \emph{per tile} of the encoder output tensors, i.e., the data volume the rANS process encodes. 
We compute file size by a worst-case upper bound by the encoder output tensor dimensionality (\Cref{subsec:comparch}) without serialization formats that could exploit the sparsity of $\boldsymbol{\hat{\mathrm{y}}}$ and $\boldsymbol{\hat{\mathrm{z}}}$. 
There are two essential findings from the results. First, the rANS process can consume tasks considerably faster than the inference process can produce them, i.e., there is no risk of backpressure within the pipeline. Second, there is only a minimal percentage decrease in TCR/s across all devices and models relative to sequential execution. Hence, we argue that the pipeline successfully offsets the runtime overhead as claimed in \Cref{sec:conctask} even without relying on a precomputed lookup table (e.g., tANS in ZSTD~\cite{collet2018zstandard}).
The results are unsurprising when viewing the CPU and GPU load of DNN inference without CPU-bound concurrent tasks in \Cref{plot:cpugpunoconc}. 
\input{embed/media/plot_cpu_gpu_usage_no_conc}
Since inference is GPU-bound, CPU usage is low even when the GPU is under maximal load. Contrast this with the CPU and GPU usage in \Cref{plot:cpugpuwithconc} where we monitor~\cite{e2ebench} usage while running the entire pipeline with the two concurrent processes.
\input{embed/media/plot_cpu_gpu_usage_with_conc}
If the CPU-bound processing task were to interfere with the DNN execution, resource usage should reveal frequent drops in GPU load. Comparing FOOL-L to S and M reveals some dependency between DNN size and CPU usage. For FOOL-L, two discernible drops in GPU usage suggest some interference, which may explain the 2.9\% decrease in TCR/s for FOOL-L and BSVBI-L. In contrast, there is no noticeable pattern difference in GPU load between \Cref{plot:cpugpuwithconc} and \Cref{plot:cpugpunoconc} for S and M variants, explaining the negligible 1-1.5\% TCR/s drop.
\subsubsection{Downlinkable Data Volume} \label{subsubsec:tilestosaturate}
We now compare how methods can alleviate the downlink bottleneck using the traces from previous experiments.
\Cref{plot:tilestosaturation} visualizes the transferable volume per 
downlink pass. 
\input{embed/media/plot_tiles_to_saturation}
Notice the logarithmic scale, i.e., FOOL improves downlinking using bent pipes by over two orders of magnitude without relying on prior information on the downstream tasks or crude filtering methods. For example, given Maxar's WorldView-3 conditions~\cite{worldview3}, it would be possible to downlink roughly 9TB of sensor data per pass before reaching downlink saturation. As a comparison, the state-of-the-art filtering method in~\cite{kodan} reports a 3$\times$ improvement based on a definition of value. 
Note that to provide a realistic presentation of the opportunities SVBI provides, we assume that a nanosatellite processes tiles until reaching a downlink segment. Moreover, we disregard the ``computational deadline'', i.e., it can process all the data before reaching a ground segment. This is reasonable since there should always be enough data to process. If not produced by a single sensor, constellations may designate certain satellites as compression nodes using reliable, high-capacity local communication channels~\cite{kepler2020}. Further, 
it is inferrable that even the low-end current-generation Orin Nano 
without any vendor-specific optimization would barely miss the computational deadline.
\subsubsection{Energy Consumption and Savings} \label{subsubsec:energycompsavings} 
%
The following investigates the energy usage of the selected model for each device. As the GPU and CPU usage patterns are highly similar,
we measure by the time it takes until a method can \emph{double} the downlinkable data. For example, if only downlinking 40 GB is possible with the unprocessed data, then we measure energy cost until the encoded size corresponds to 80 GB of raw captures. \Cref{plot:energyprocessing} summarizes the results.
\input{embed/media/plot_energy_cost_processing}
As expected, processing on TX2 requires more energy than on NX and Orin Nano as it is slower. Somewhat interesting is that the NX consumes more energy than the Orin Nano. As they execute the same models with comparable processing throughput, the results suggest that the newer Jetson lineup is more energy-efficient. 
Lastly, we measure savings from reduced transmission time, arguably an often undervalued advantage of compression. Admittedly, satellites will downlink as bandwidth permits, i.e., the transmission energy cost does not depend on the codec performance when there is saturation. Nonetheless, to intuitively show the amount of energy large volumes might require, we contrast with bent pipes in \Cref{plot:transmissionsavings}. 
\input{embed/media/plot_energy_transmission_savings}
Given the link conditions, we measure the difference in energy cost between transmitting until saturation and transmitting the corresponding raw volume from \Cref{plot:tilestosaturation}.
%

%% file: embed/media/plot_throughput_main.tex
\begin{figure}[htb]
    \centering
    \includegraphics[width=\columnwidth]{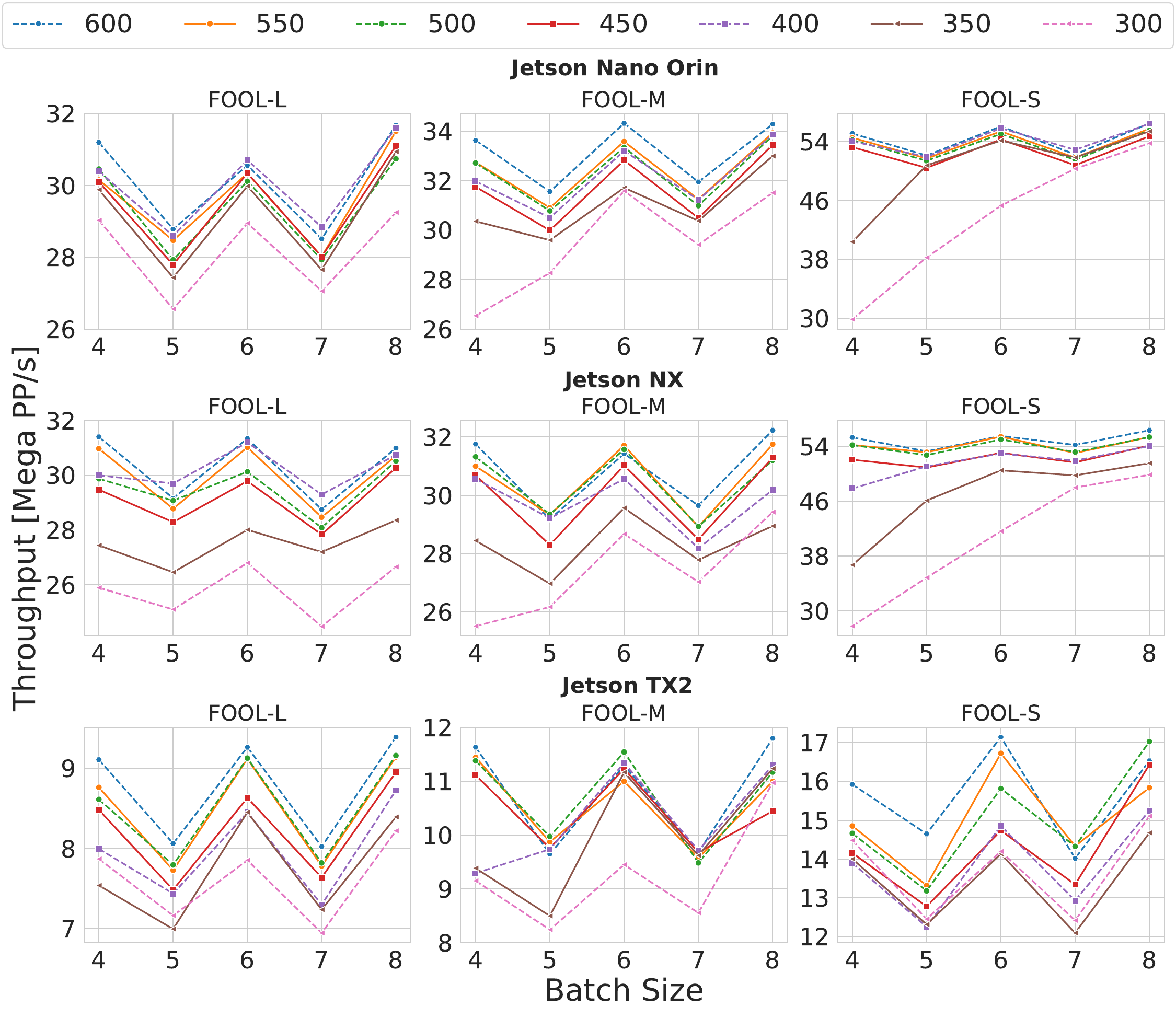}
    \caption{Processing Throughput by Model Size}
    \label{plot:throughputmain}
\end{figure}

%% file: embed/tables/optim_config_memory_wthroughput.tex
\begin{table}[htb]
\centering
\caption{Throughput Comparison Between Feature Codecs}
\label{tab:optimconfigmemorywthroughput}
\begin{tabularx}{\columnwidth}{@{\extracolsep{\fill}} 
c
l >{\centering\arraybackslash}p{0.18\linewidth}@{ }@{ }
  >{\centering\arraybackslash}p{0.15\linewidth}@{ }@{ }
r }
Device &   Model &  Spatial Dim. & 
  Batch Size &  TCR/s \textuparrow\textuparrow \\ \toprule
 & FOOL-L & 600x600 & 4 & $7.4794\cdot10^{8}$ \\
 & \cellcolor[HTML]{EFEFEF}FOOL-M &
  \cellcolor[HTML]{EFEFEF}600x600 &
  \cellcolor[HTML]{EFEFEF}8 &
  \cellcolor[HTML]{EFEFEF}$7.6694\cdot10^{8}$ \\
\multirow{-3}{*}{Orin Nano} &
  \textbf{FOOL-S} &  600x600 &  8 &
  $\boldsymbol{1.3386\cdot10^{9}}$ \\ \hline
 & FOOL-L &  600x600 &  8 &  $7.5456\cdot10^{8}$ \\
 &\cellcolor[HTML]{EFEFEF}FOOL-M &
  \cellcolor[HTML]{EFEFEF}600x600 &
  \cellcolor[HTML]{EFEFEF}6 &
  \cellcolor[HTML]{EFEFEF}$8.1664\cdot10^{8}$ \\
\multirow{-3}{*}{NX} &  \textbf{FOOL-S} &  600x600 &  8 &
  $\boldsymbol{1.3422\cdot10^{9}}$ \\ \hline
 &\cellcolor[HTML]{EFEFEF}FOOL-L &
  \cellcolor[HTML]{EFEFEF}600x600 &
  \cellcolor[HTML]{EFEFEF}8 &
  \cellcolor[HTML]{EFEFEF}$2.3696\cdot10^{8}$ \\
 & FOOL-M &  600x600 &  8 &  $2.8067\cdot10^{8}$ \\
\multirow{-3}{*}{TX2} &
  \cellcolor[HTML]{EFEFEF}\textbf{FOOL-S} &
  \cellcolor[HTML]{EFEFEF}600x600 &
  \cellcolor[HTML]{EFEFEF}6 &
  \cellcolor[HTML]{EFEFEF}$\boldsymbol{4.0751\cdot10^{8}}$ \\ \hline
 &BSVBI-L &  600x600 &  8 &  $6.1419\cdot10^{8}$ \\
 &\cellcolor[HTML]{EFEFEF}\textbf{BSVBI-M} &
  \cellcolor[HTML]{EFEFEF}600x600 &
  \cellcolor[HTML]{EFEFEF}5 &
  \cellcolor[HTML]{EFEFEF}$7.2504\cdot10^{8}$ \\
\multirow{-3}{*}{Orin Nano} &
  BSVBI-S &  550x550 &  7 &  $6.9919\cdot10^{9}$ \\ \hline
 &\cellcolor[HTML]{EFEFEF}BSVBI-L &
  \cellcolor[HTML]{EFEFEF}600x600 &
  \cellcolor[HTML]{EFEFEF}8 &
  \cellcolor[HTML]{EFEFEF}$6.0007\cdot10^{8}$ \\
 &\textbf{BSVBI-M} &  600x600 &  6 &  $7.3717\cdot10^{8}$ \\
\multirow{-3}{*}{NX} &
  \cellcolor[HTML]{EFEFEF}BSVBI-S &
  \cellcolor[HTML]{EFEFEF}600x600 &
  \cellcolor[HTML]{EFEFEF}6 &
  \cellcolor[HTML]{EFEFEF}$7.0720\cdot10^{9}$ \\ \hline
 &BSVBI-L &  600x600 &  7 &  $1.7861\cdot10^{8}$ \\
 &\cellcolor[HTML]{EFEFEF}\textbf{BSVBI-M} &
  \cellcolor[HTML]{EFEFEF}600x600 &
  \cellcolor[HTML]{EFEFEF}6 &
  \cellcolor[HTML]{EFEFEF}$2.9978\cdot10^{8}$ \\
\multirow{-3}{*}{TX2} &  BSVBI-S &  600x600 &  7 &  $2.6818\cdot10^{8}$ \\ \bottomrule 
\end{tabularx}
\end{table}

%% file: embed/tables/entropycoderperfwbaselines.tex
\begin{table}[htb]
\centering
\caption{Concurrent Entropy Coding and Effect on TCR/s}
\label{tab:entropycodingfiles}
\begin{tabularx}{\columnwidth}{@{\extracolsep{\fill}} 
l
>{\raggedleft\arraybackslash} p{0.14\linewidth}
r
>{\raggedleft\arraybackslash} p{0.118\linewidth}
>{\centering\arraybackslash} p{0.09\linewidth}
>{\raggedleft\arraybackslash} p{0.162\linewidth} }
Model & \makecell[c]{TCR/s\\$[$conc$]$} & \makecell[c]{TCR/s\\dec.} &
\makecell[c]{File Size\\(MB)}
&  \makecell[c]{File/s} &
\makecell[c]{rANS conc.\\(MB/s)}
\\ 
\toprule 
\rowcolor[HTML]{EFEFEF} 
FOOL-L  & $7.26 \cdot 10^{8}$ & \textbf{2.94\%} & 0.616 & 29 & 37.2                         \\
FOOL-M  & $7.57 \cdot 10^{8}$ & 1.28\%          & 0.462 & 31 & 38.3                         \\
\rowcolor[HTML]{EFEFEF} 
FOOL-S &
  \cellcolor[HTML]{EFEFEF}$1.32 \cdot 10^{9}$ &
  \cellcolor[HTML]{EFEFEF}1.06\% &
  0.383 &
  52 &
  38.8 \\ \hline
BSVBI-L & $5.96 \cdot 10^{8}$ & \textbf{2.88\%} & 0.822 & 37 & 37.6                         \\
\rowcolor[HTML]{EFEFEF} 
BSVBI-M & $7.15 \cdot 10^{8}$ & 1.37\%          & 0.617 & 42 & \cellcolor[HTML]{EFEFEF}39.6 \\
BSVBI-S & $6.91 \cdot 10^{8}$ & 1.28\%          & 0.437 & 58 & 40.1                         \\
\bottomrule 
\end{tabularx}%
\end{table}

%% file: embed/media/plot_cpu_gpu_usage_no_conc.tex
\begin{figure}[htb]
    \centering
    \includegraphics[width=\columnwidth]{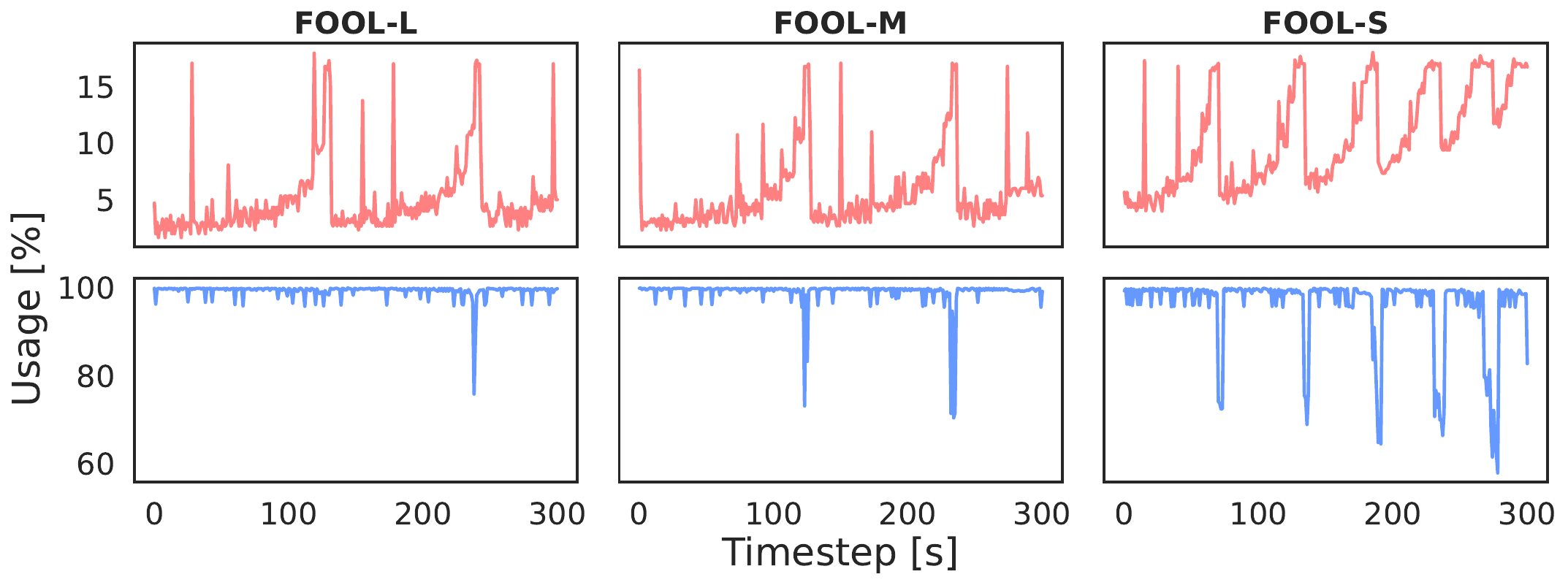}
    \caption{CPU (red) and GPU (blue) Usage of Encoder Network}
    \label{plot:cpugpunoconc}
\end{figure}

%% file: embed/media/plot_cpu_gpu_usage_with_conc.tex
\begin{figure}[htb]
    \centering
    \includegraphics[width=\columnwidth]{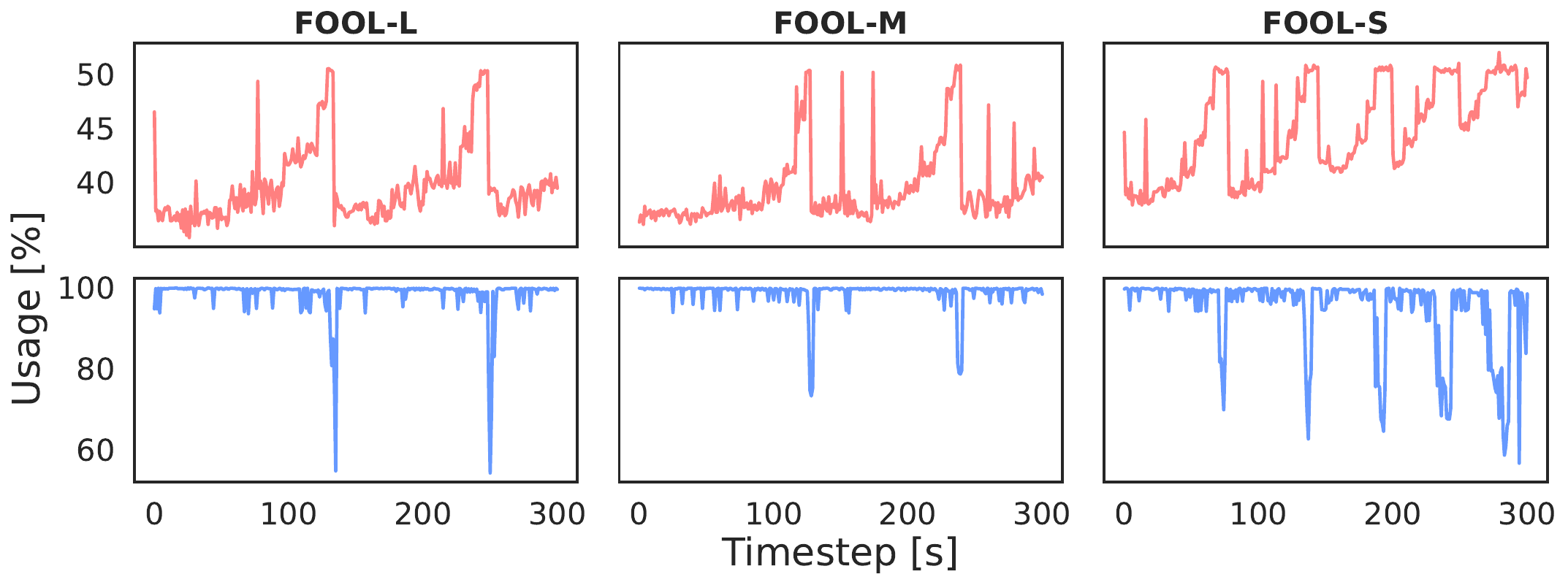}
    \caption{CPU (red) and GPU (blue) Usage of Concurrent Pipeline}
    \label{plot:cpugpuwithconc}
\end{figure}

%% file: embed/media/plot_tiles_to_saturation.tex
\begin{figure}[htb]
    \centering
    \includegraphics[width=\columnwidth]{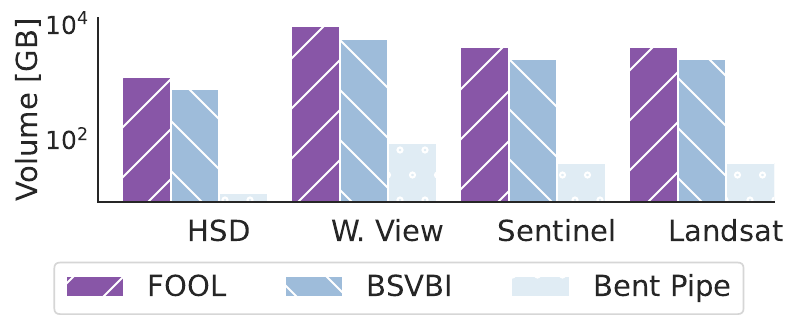}
    \caption{Downlinkable Data Volumes by Link}
    \label{plot:tilestosaturation}
\end{figure}

%% file: embed/media/plot_energy_cost_processing.tex
\begin{figure}[htb]
    \centering
    \includegraphics[width=\columnwidth]{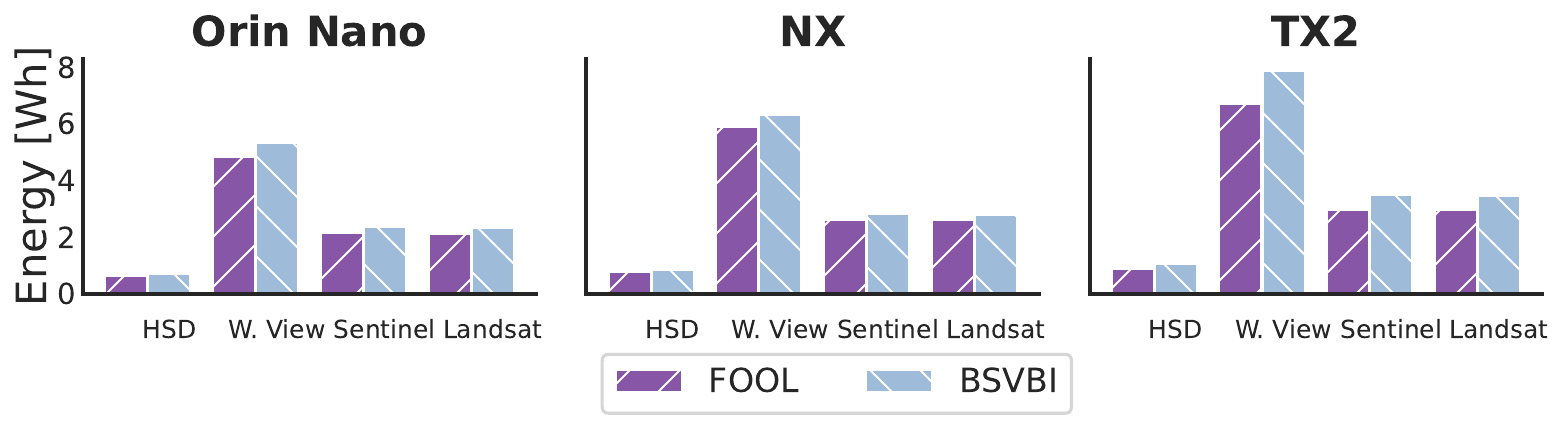}
    \caption{Energy Cost of Compression Pipelines}
    \label{plot:energyprocessing}
\end{figure}

%% file: embed/media/plot_energy_transmission_savings.tex
\begin{figure}[htb]
    \centering
    \includegraphics[width=\columnwidth]{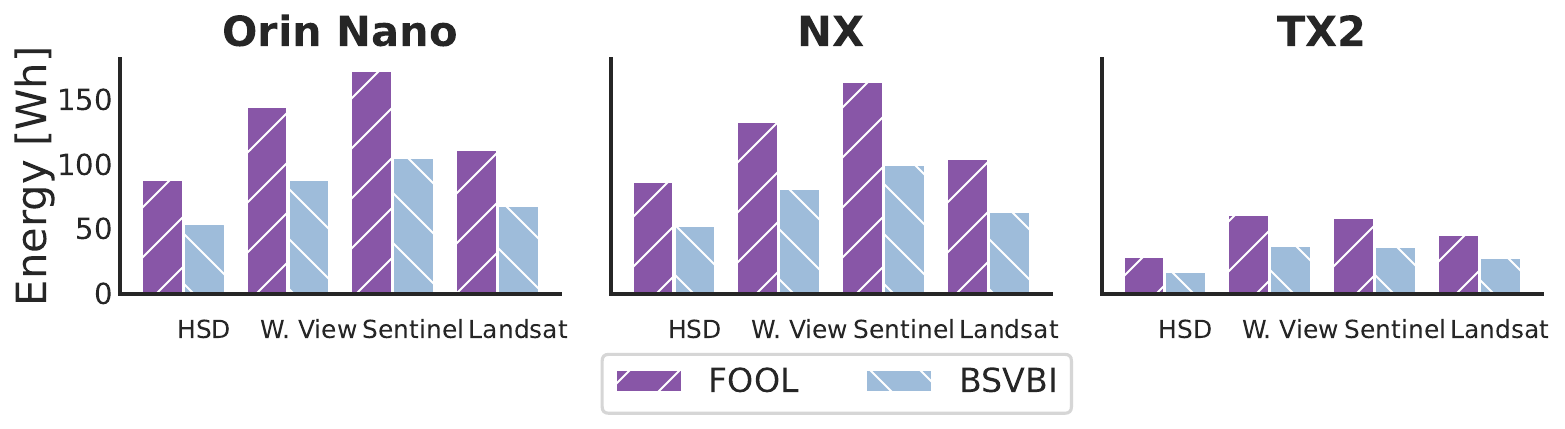}
    \caption{Potential Energy Savings from Transmission}
    \label{plot:transmissionsavings}
\end{figure}

%% file: sections/limitations.tex
\section{Discussions and Limitations} \label{sec:disclim}
\subsection{Downlink Saturation Handling} \label{subsec:sathandling}
This work has omitted to handle downlink saturation, i.e., prioritizing salient data when compression cannot sufficiently reduce the volume in time. Existing work suggests to apply intelligent adhoc filtering. However, as argued in \Cref{sec:relwork}, such filtering relies on strong assumptions that bias the downlinked data towards a small subset of tasks clients may be interested in. Therefore, we suggest that filtering should prioritize tiles that provide sufficient information to recover filtered tiles with a generative model.
%
Intuitively, the lower the remaining uncertainty, given other tiles, the more reliable generative models can recover missing tiles.
\subsection{Reliance on Foundational Models}
Earth Observation (EO) requires considerations not included in common object detection objectives and architectural components, and widespread foundational models for satellite imagery have yet to emerge. Nonetheless, proprietary offerings already exist~\cite{deci2024}, and we argue that the community drives to open-source solutions will inevitably mitigate the limitation.
We worked around not having access to an EO-native foundational model by only freezing the shallow layers to train the predictors, i.e, each predictor is complemented with a backbone suitable for a particular sensor configuration.

%% file: sections/conclusion.tex
\section{Conclusions} \label{sec:conclusion}
This work introduced a novel compression method that addresses the downlink bottleneck in LEO without relying on prior knowledge of downstream tasks. A rigorous evaluation showed that FOOL increases data volume with advancements that, to the best of our knowledge, are unprecedented.
The rate reductions are primarily from the task-agnostic context. Additionally, the ITA layers further improve compression performance with an overhead that does not outweigh the processing throughput gains from batch parallelization. Lastly, we transparently listed limitations that future work should consider and identified promising future research directions for OEC based on novel insights.

%% file: biography/biography.tex
\vspace{-15mm}
\begin{IEEEbiography}[{\includegraphics[width=1in,height=1.25in,clip,keepaspectratio]{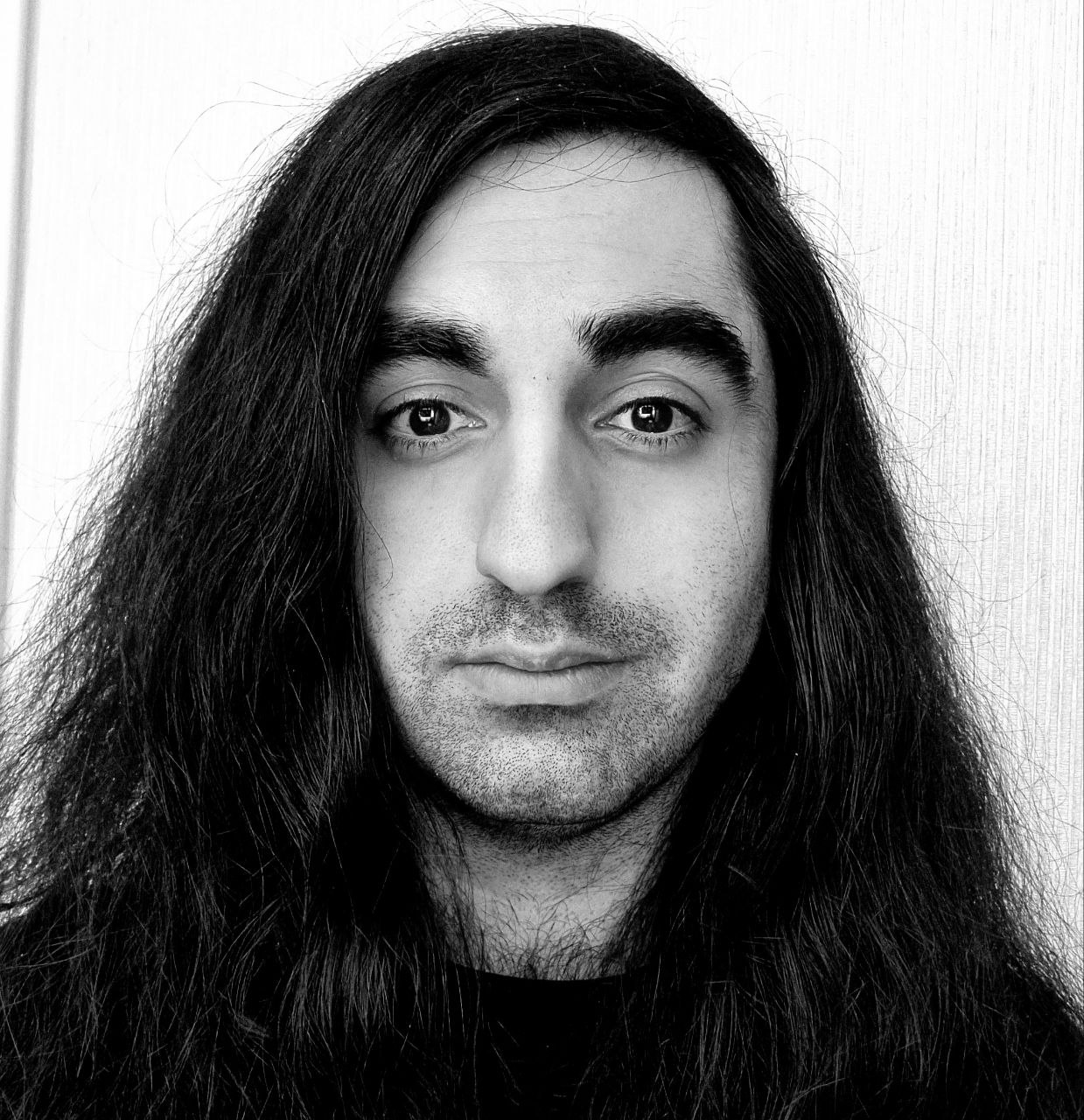}}]{Alireza Furutanpey} received a MSc from the Technical University of Vienna, Austria in 2022 with distinction in the field of Computer Science. He is now a PhD candidate at the Distributed Systems Group in the field of Edge Computing. His research interests include Mobile Edge Computing, Edge Intelligence and Machine Learning.
\end{IEEEbiography}
\vspace{-15mm}
\begin{IEEEbiography}[{\includegraphics[width=1in,height=1.25in,clip,keepaspectratio]{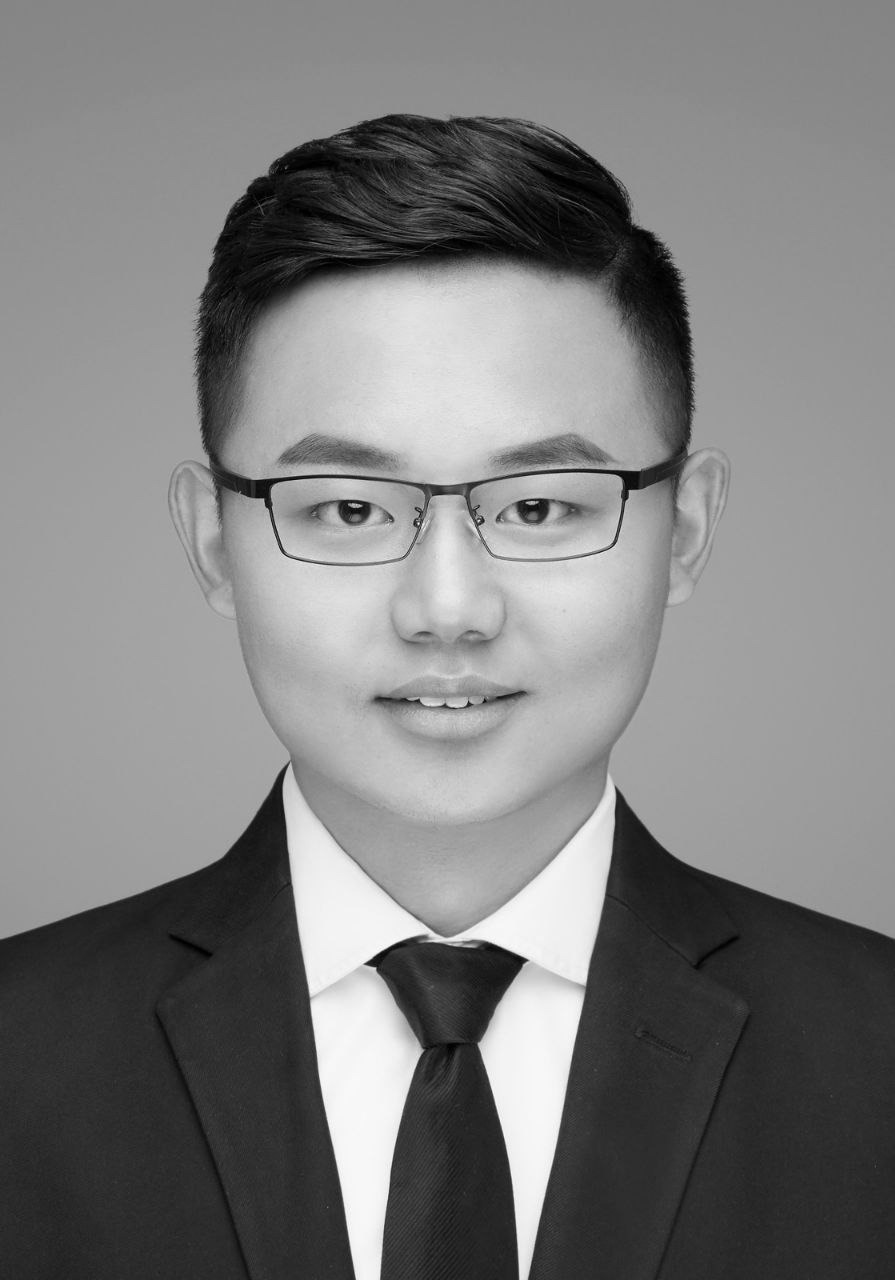}}]{Qiyang Zhang} is a Ph.D. candidate in computer science at the State Key Laboratory of Networking and Switching Technology, Beijing University of Posts and Telecommunications. He was also a visiting researcher at the Distributed Systems Group at TU Wien from December 2022 to December 2023. His research interests include Satellite Computing, Edge Intelligence
\end{IEEEbiography}
\vspace{-15mm}
\begin{IEEEbiography}[{\includegraphics[width=1in,height=1.25in,clip,keepaspectratio]{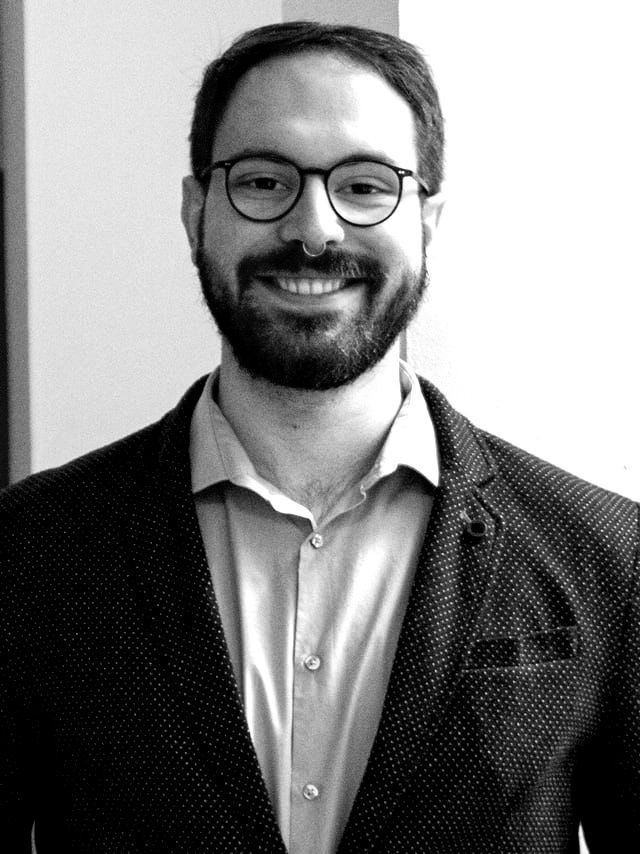}}]{Philipp Raith} received a MSc from the Technical University of Vienna, Austria in 2021 with distinction in the field of Computer Science. He is now a PhD candidate at the Distributed Systems Group in the field of Edge Computing. His research interests include Serverless Edge Computing, Edge Intelligence and Operations for AI.
\end{IEEEbiography}
\vspace{-15mm}
\begin{IEEEbiography}[{\includegraphics[width=1in,height=1.25in,clip,keepaspectratio]{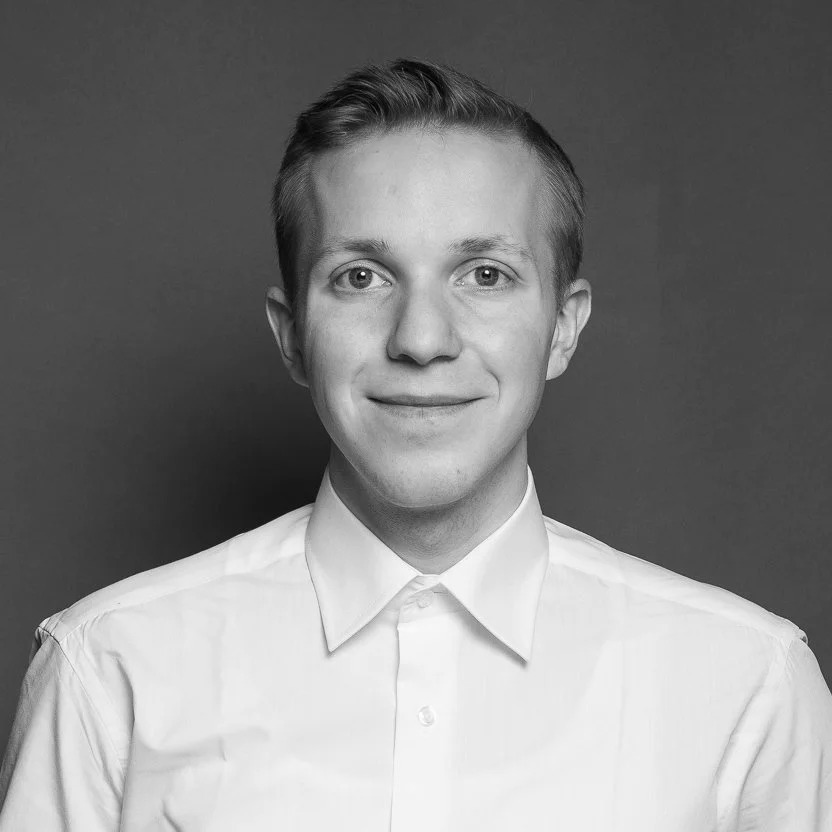}}]{Tobias Pfandzelter} has been a research associate and PhD student at the Scalable Software Systems research group since September 2019. Before that, he completed his Bachelor and Master in Computer Science at TU Berlin. His research focus is on edge computing in LEO satellite constellations.
\end{IEEEbiography}
\vspace{-15mm}
\begin{IEEEbiography}[{\includegraphics[width=1in,height=1.25in,clip,keepaspectratio]{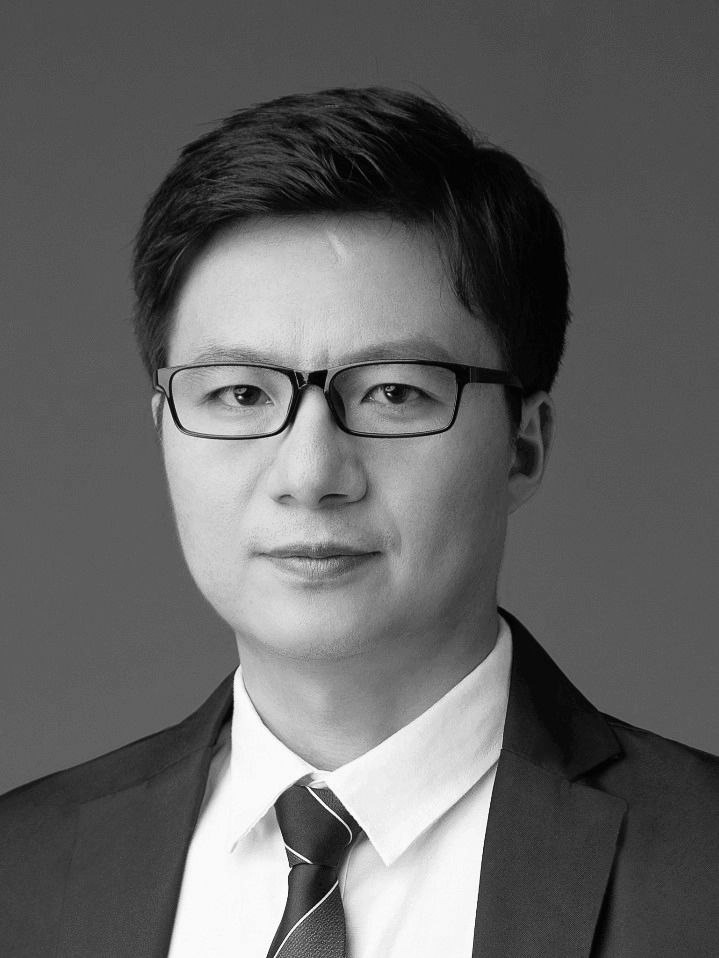}}]{Shangguang Wang} is a professor at the School of Computer Science, Beijing University of Posts and Telecommunications, China.
His research interests include service computing, mobile edge computing, cloud computing, and satellite computing. He is currently serving as chair of IEEE Technical Community on Services Computing(TCSVC), and vice chair of IEEE Technical Community on Cloud Computing. He also served as general chairs or program chairs of 10+ IEEE conferences, advisor/associate editors of several journals
such as Journal of Cloud Computing, Journal of Software: Practice and Experience, International Journal of Web and Grid Services, China Communications, and so on. He is a senior member of the IEEE, and Fellow of the IET.
\end{IEEEbiography}
\vspace{-14mm}
\begin{IEEEbiography}[{\includegraphics[width=1in,height=1.25in,clip,keepaspectratio]{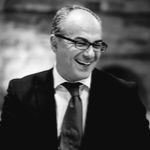}}]{Schahram Dustdar} is a full professor of computer science and heads TU Wien's Distributed Systems Group. His research interests include distributed systems, Edge Intelligence, complex and autonomic software systems. He's the editor in chief of Computing;  associate editor of ACM Transactions on the Web, ACM Transactions on Internet Technology, IEEE Transactions on Cloud Computing, and IEEE Transactions on Services Computing. He's also on the editorial boards of IEEE Internet Computing and IEEE Computer. He has received the ACM Distinguished Scientist award and Distinguished Speaker Award and the IBM Faculty Award. He is an elected member of Academia Europaea, where he's was Informatics Section chairman from 2015 to 2022. He is an IEEE Fellow and AAIA Fellow where he is the current President.
\end{IEEEbiography}